%% file: main.tex
\documentclass[twoside]{article}

\usepackage[accepted]{aistats2020}
\usepackage[round]{natbib}
\renewcommand{\bibname}{References}
\renewcommand{\bibsection}{\subsubsection*{\bibname}}

\input{math_commands.tex}

\usepackage[hide=true,setmargin=true,marginparwidth=.6in]{marginalia}
\input{defs.tex}
\usepackage{hyperref}
\usepackage{url}
\usepackage{graphicx}
\usepackage[capitalize]{cleveref}
\usepackage{amsthm,thmtools}
\DeclareMathAlphabet\EuRoman{U}{eur}{m}{n}
\SetMathAlphabet\EuRoman{bold}{U}{eur}{b}{n}

\usepackage{multirow}
\usepackage{tabularx}
\newcolumntype{Y}{>{\centering\arraybackslash}X}

\crefname{lemma}{Lemma}{Lemmas}
\crefname{corollary}{Corollary}{Corollaries}
\crefname{theorem}{Theorem}{Theorems}

\makeatletter
\let\reftagform@=\tagform@
\def\tagform@#1{\maketag@@@{\ignorespaces\textcolor{gray}{(#1)}\unskip\@@italiccorr}}
\renewcommand{\eqref}[1]{\textup{\reftagform@{\ref{#1}}}}
\makeatother

\declaretheorem[style=plain,numberwithin=section,name=Theorem]{theorem}

\declaretheorem[style=plain,sibling=theorem,name=Proposition]{proposition}

\declaretheorem[style=definition,sibling=theorem,name=Definition]{definition}
\numberwithin{theorem}{section}

\makeatletter
\newcommand{\printfnsymbol}[1]{%
  \textsuperscript{1}%
}
\makeatother

\renewcommand*{\thefootnote}{\fnsymbol{footnote}}

\begin{document}

% If your paper is accepted and the title of your paper is very long,
% the style will print as headings an error message. Use the following
% command to supply a shorter title of your paper so that it can be
% used as headings.
%
\runningtitle{RelatIF}

% If your paper is accepted and the number of authors is large, the
% style will print as headings an error message. Use the following
% command to supply a shorter version of the authors names so that
% they can be used as headings (for example, use only the surnames)
%
\runningauthor{Barshan\textsuperscript{*}, Brunet\textsuperscript{*}, Dziugaite}

\twocolumn[

\aistatstitle{RelatIF: Identifying Explanatory Training Examples\\ via Relative Influence}
%\aistatsauthor{ Anonymous }
%\aistatsaddress{}]
% \aistatsauthor{ Elnaz Barshan\footnotemark \And Marc-Etienne Brunet\printfnsymbol{1} \And  Gintare Karolina Dziugaite}
\aistatsauthor{ Elnaz Barshan\footnotemark \And Marc-Etienne Brunet\footnotemark[1] \And  Gintare Karolina Dziugaite}
% We should re-define footnotemark to print * instead of number
\aistatsaddress{ Element AI \And  Element AI \And Element AI }
]
% \thanks{These authors contributed equally. Names are sorted alphabetically.}
\begin{abstract}
In this work, we focus on the use of influence functions to identify relevant training examples
that one might hope ``explain'' the predictions of a machine learning model.
One shortcoming of influence functions is that 
the training examples deemed most ``influential'' are often outliers or mislabelled,
making them poor choices for explanation.
In order to address this shortcoming, 
we separate the role of global versus local influence.
We introduce \RelIF{}, a new class of criteria for choosing relevant training examples by way of an optimization objective that places a constraint on global influence.
\RelIF{} considers the local influence that an explanatory example has on a prediction relative to its global effects on the model.
In empirical evaluations, we find that the examples returned by RelatIF are more intuitive when compared to those found using influence functions.
\end{abstract}
%----------------------------------------------------------------------
\section{Introduction}
%----------------------------------------------------------------------
% \TBD{
% \begin{itemize}
%     \item reliability and trustworthiness of the ML algorithm predictions
%     \item ML models can be fooled by adding imperceptible change to the input
%     \item potential for unreasonable behaviour => explainability is required.
%     \item make sure ML models reflects our values
%     \item explaining the reasoning process of the model is sometimes a legal requirement[GRPR].
%     \item limited use of black-box models in fields with high-regulatory requirements.
%     \item providing exp to end users 
% \end{itemize}
% }
% Despite the high prediction accuracy of modern machine learning models, they have the potential to behave unreasonably~\citep{szegedy2013intriguing, goodfellow2014explaining, nguyen2015deep}.
% Because of this, 
As the use of black-box models becomes widespread, many believe that their predictions  
% Given this potential and the widespread use of black-box machine learning models, model predictions 
must be accompanied by interpretable explanations~\citep{lipton2016mythos, doshi2017towards, goodman2017european}. % ADD GDPR
There is a growing body of research concerned with how to best explain black-box predictions ~\citep{kim2017interpretability,ribeiro2018anchors,lundberg2017unified}.
% One way that experts explain their predictions is by supplying supporting examples. %specific examples that support their predictions.
% These examples may be confirmatory, showcasing a similar situation where the prediction would have been appropriate, 
% or they may provide contrast, helping to rule out alternative predictions.
% Examples can ground a complex prediction in concrete data and help establish confidence in expertise. 
% In the machine learning context, 
A natural way to explain a model prediction is to provide supporting examples from the training data.
However, one must \textit{choose} which examples are most relevant.
One approach is to trace the model prediction back to the training examples and determine the impact of each individual training example. %and choose those having the greatest impact on that prediction.
% Indeed, for a machine, it is straightforward to determine how a prediction changes if, say, the role of particular training example is up/down-weighed.
This idea underlies so-called \emph{influence functions} (IF), 
which quantify the impact of an individual example on a prediction in a differentiable model \citep{jaeckel1972infinitesimal, hampel1974influence}.
The use of IF to \textit{explain} black-box predictions by training examples with high influence was pioneered by \citet{koh2017understanding}.
%The use of IF to \textit{explain} black-box predictions was pioneered by \citet{koh2017understanding},
%where they employ IF to identify training examples that are most ``responsible'' for the generated prediction.

\begin{figure*}[!ht]
	\centering
	\begin{tabular}{c c c c}
		\includegraphics[width=0.28\linewidth]{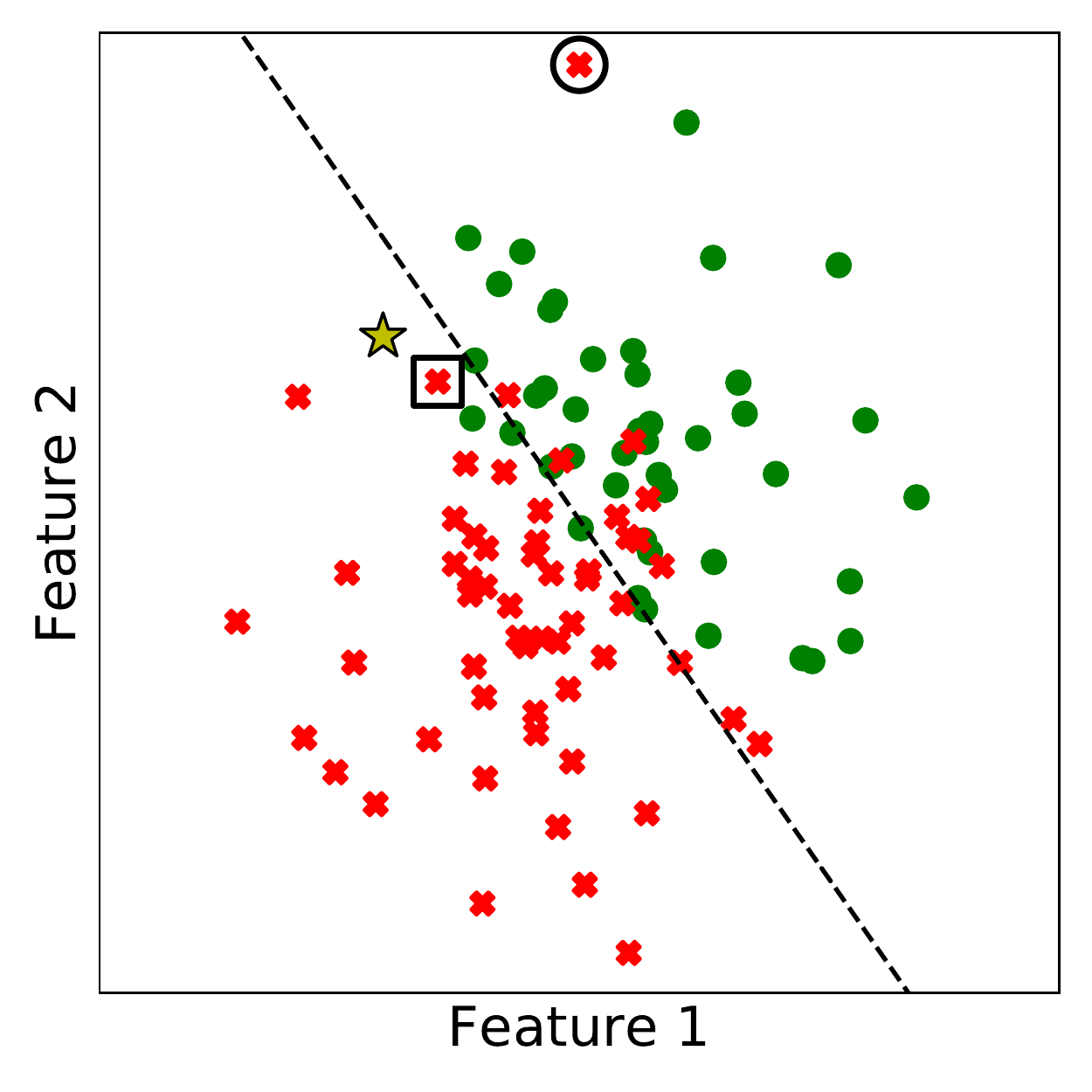} & ~~~~~ &
	    \includegraphics[width=0.28\linewidth]{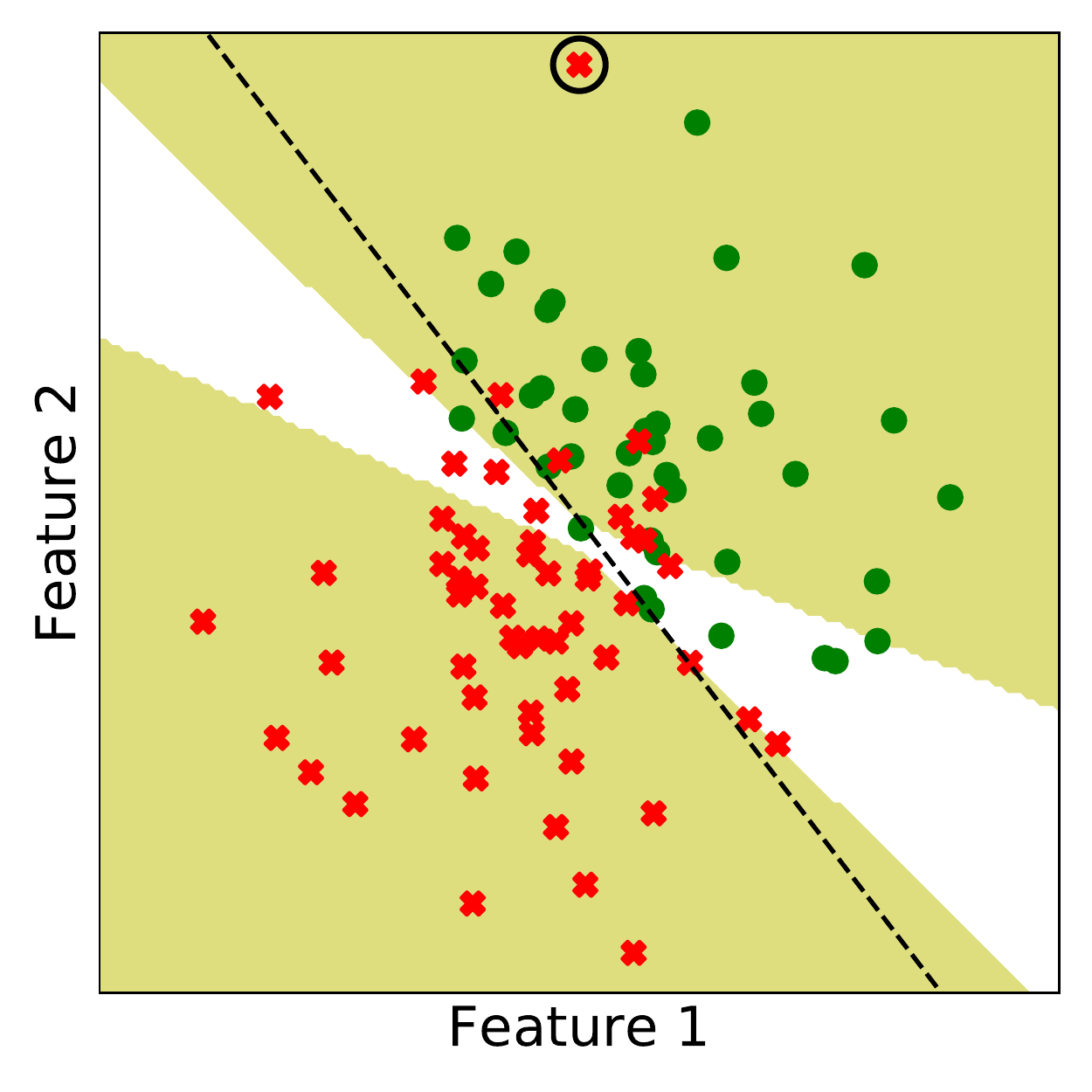} &
	    \includegraphics[width=0.28\linewidth]{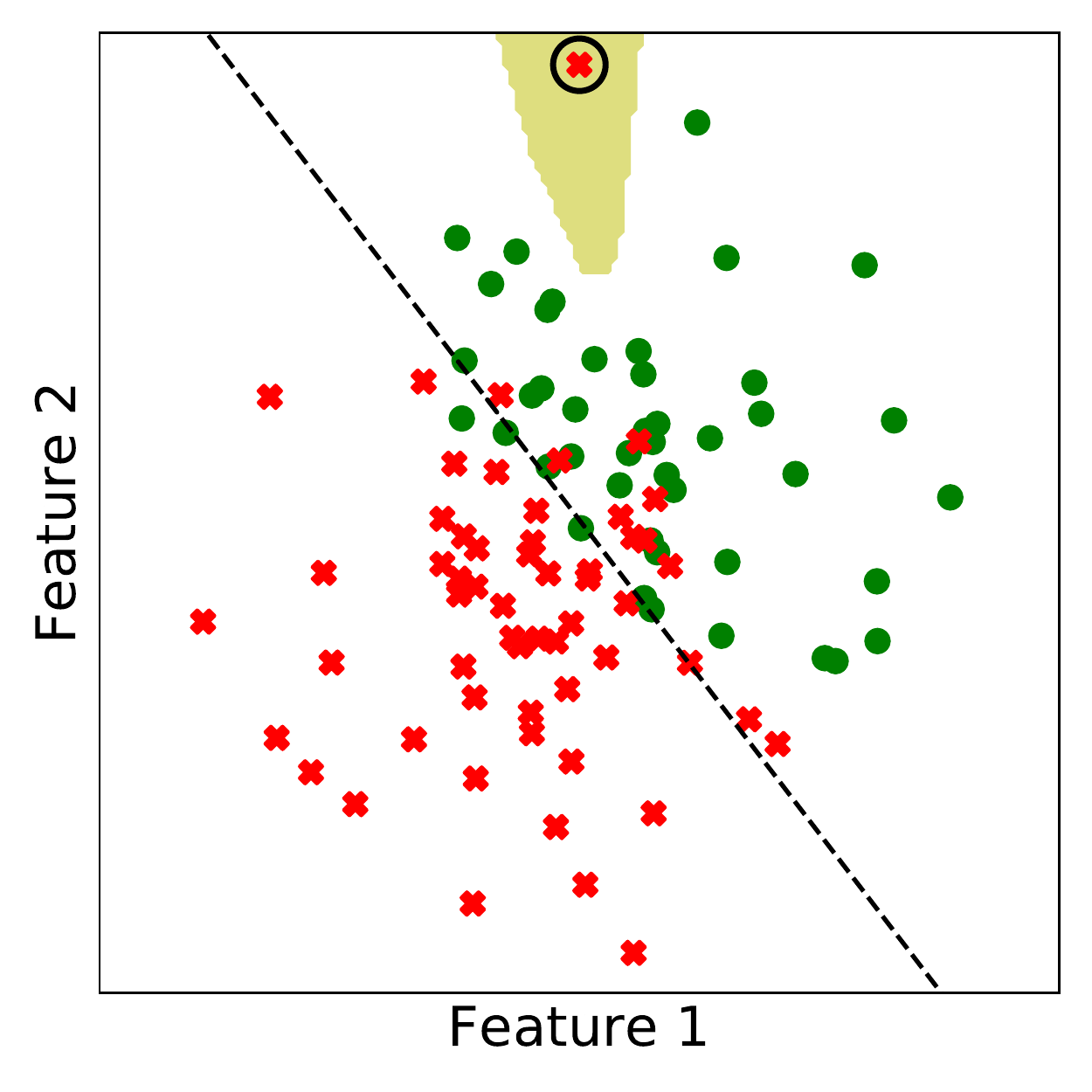} \\
	    (a) & & (b) & (c)
	\end{tabular}
	\caption{
	    Binary classification by linear decision boundaries (dashed line) to illustrate the difference between IF and \RelIF{}. 
	    (a) The model makes a prediction for the test input (star). 
	    As estimated by IF, the most influential training example for the prediction is an outlier (circled). 
	    Using \RelIF{}, the most influential training example is more \NA{typical} (encased in a square). 
	    (b) Using IF, every test input falling within the shaded yellow region is most influenced by the same outlier (circled). Test inputs in the remaining white region are most influenced by one of 5 other high loss examples.
	    (c) Using \RelIF{}, the region where test inputs are most influenced by the outlier (circled) shrinks. Test inputs in the remaining region are most influenced by one of 65 other examples.}
	\label{fig:illustrative}
\end{figure*}

In this work, we revisit the use of IF for identifying examples to explain predictions.
Our motivation is the observation that the most influential training examples returned by IF are often mislabelled or outliers, 
on the account of these examples also incurring high loss (see \cref{fig:illustrative}(a)).
%or otherwise high loss (see \cref{fig:illustrative}(a)), 
This phenomenon is well documented in the literature~\citep{hampel1974influence, campbell1978influence, cook1982residuals, croux2000principal}.
As a result, we find that the set of examples deemed ``influential'' tend to be highly overlapping, in the sense that the predictions for many different inputs are all most affected by the same small set of training examples.
In other words, these training examples seem to have a global impact on the model.
\cref{fig:illustrative} (b) illustrates these effects. 
Returning examples from an overlapping set of high loss training examples may be desirable when the recipient of the explanation is attempting to debug or otherwise improve the model.
However, these examples can be unsatisfying as an explanation for an end user.
Explaining a prediction by returning high loss training examples, which are often unusual or misclassified, does not inspire confidence in the model. 
Nor does it inspire confidence when, for several seemingly different inputs, the same high-loss examples are returned as an explanation. 
% We think it is preferable to explain a prediction through more typical examples, selected in a way that is more specific to the model input.\fTBD{GKD: We do not \emph{seek} typical examples. The examples returned by normalized IF also happen to be typical ones.}

\footnotetext{Equal contribution, alphabetic order}
\renewcommand*{\thefootnote}{\arabic{footnote}}
\setcounter{footnote}{0}

We propose \RelIF{}, a new class of criteria for identifying ``relevant'' training examples to explain a prediction.
\RelIF{} considers the local influence that an explanatory example has on a prediction relative to its global effects on the model.
We formulate these criteria in terms of a constrained optimization problem:
for a given allowable global ``impact'' on the model,
which training examples most influence the specific prediction?
We argue that this approach is better suited to producing an explanatory training example for an end user.
\cref{fig:illustrative}(c) illustrates how this objective dampens the influence of an outlier.
We show experimentally that \RelIF{} returns training examples that are more typical as well as more specific to a given input when compared to those identified using IF. 
%----------------------------------------------------------------------
\section{Preliminaries}
%----------------------------------------------------------------------
% Consider a machine learning model with parameters $\theta \in \Reals^d$ that seeks to minimize the empirical average of a loss function $\loss(z,\theta)$ over a set of training examples $S=\{z_1, z_2,...,z_n\}$. 
Consider a family of predictors, 
indexed by parameter values $\theta \in \Reals^d$, each a map taking input values $x \in \mathcal{X}$ to target values $y \in \mathcal{Y}$.
Let $S=(z_1, z_2,\ldots,z_n)$ denote the training examples, where  $z_i=(x_i,y_i)\in \mathcal{X}\times\mathcal{Y}$ is a pair of an input and target. 
Let $\loss(z,\theta)$ be the loss for a point $z$ under model parameters $\theta$.
Our focus is on algorithms, like stochastic gradient descent (SGD), that 
return some first-order stationary point 
$\theta(w_1,\dots,w_n)$
of the objective
\begin{equation}
\label{eq:theta_star}
    %\thetaopt = \arg \min_{\theta} 
    \sum_{i=1}^{n} w_i \loss(z_i,\theta),
\end{equation}
i.e., 
we are studying the parameters learned by weighted empirical risk minimization, where
each $w_i$ is the weight given to the training example $z_i$.
We are primarily interested in the parameter 
$\thetaopt=\theta(1/n,\dots,1/n)$ obtained by weighting all points equally.
Given an unseen input $x$, 
our goal is to identify training examples $z_i$
that, through training, 
had a ``meaningful'' influence on the prediction that $\thetaopt$ renders for $x$.

A simple way to measure the effect of each example $z_i$ on training is to re-weight its contribution, retrain from scratch, and evaluate the change to the learned parameters.
To that end, let
$\thetaepsi=\theta(1/n,\dots,1/n+ \epsilon_i,\dots,1/n)$,
where only the $i$'th weight is modified to be $w_i = \frac 1 n + \epsilon$.
Any deviations of $w_i$ from $\frac 1 n $ is denoted by $\epsilon_i$, i.e., $\epsilon_i = w_i - \frac 1 n$.
Taking $\epsilon_i = -\frac 1 n$, equivalently, setting $w_i=0$, constitutes dropping an example from the training set.

Exhaustive retraining, however, 
is computationally prohibitive. 
Instead, we will make an infinitesimal analysis,
as was pioneered in the development of \emph{influence functions} \citep{hampel1974influence,jaeckel1972infinitesimal}.
To do so, we will assume that the parameter $\theta(w_1,\dots,w_n)$ is a differentiable function of the weights at $w_1 = \dotsm = w_n = \frac 1 n$. Further, we will assume that $\loss(z,\theta)$ is strictly convex, and twice continuously differentiable, in $\theta$.

The method of \emph{influence functions} (IF) allows us to approximate how (some function of) the learned parameters $\thetaopt$ would change if we were to reweight a training example $z_i$.
The key idea is to make a first-order approximation of change in $\thetaopt$ around $\epsilon_i = 0$.
The same method can be extended to approximate changes in any twice continuously differentiable functions $f$ of the parameters around $\epsilon_i = 0$. 
Namely, one can approximate how some $f(\thetaepsi)$ changes with $\epsilon_i$ by considering $\dee f(\thetaepsi) / \dee \epsilon_i \rvert_{\epsilon_i = 0}$ .
%Restated \fTBD{GKD: Not sure I get this, but may be correct.}, IF lets us approximate how some $f(\thetaepsi)$ changes with $\epsilon_i$ by considering $\dee f(\thetaepsi) / \dee \epsilon_i \rvert_{\epsilon_i = 0}$ .
Indeed, our primary interest is the influence on the loss for a test sample.

\begin{definition}
\label{def:influence}
Fix a test sample $z_\testpnt = (x_\testpnt, y_\testpnt)$.
The \emph{influence of $z_i$ on (the loss of) $z_{\testpnt}$} is defined to be
\begin{align*}
    \Infl{\testpnt}{i} = -\frac{\dee \loss(z_{\testpnt},\thetaepsi)}{\dee \epsilon_i}\bigg\rvert_{\epsilon_i = 0} 
        = -g^T_{\testpnt} \frac{\dee \thetaepsi}{\dee \epsilon_i}\bigg\rvert_{\epsilon_i = 0},
\end{align*}
where $\gradient{\testpnt} = \grad_{\thetaopt} \loss(z_\testpnt,\thetaopt)$.
\end{definition}

Note that the second equality follows from the chain rule 
and that $\thetaepsi \rvert_{\epsilon_i = 0} = \thetaopt$.
%We refer to $\Infl{\testpnt}{i}$  as simply \emph{the influence of $z_i$ on $z_\testpnt$}, when the context is clear.

Because we consider a negative change in \cref{def:influence}, a training example $z_i$ having \emph{positive} influence on $z_\testpnt$ \emph{decreases} the loss $\loss(z_\testpnt, \thetaepsi)$ when upweighted and may thus be considered helpful.
A training example having \textit{negative} influence on $z_\testpnt$ \emph{increases} the loss and may thus be considered harmful. 

% We are often interested in situations where the model prediction $y_\testpred$ does not agree with the target $y_\testpnt$, or where we do not have access to the ground truth target at all. 
% In these cases, we may consider the influence of $z_i$ on $z_\testpred = (x_\testpnt, y_\testpred)$, which we denote $\Infl{\testpred}{i}$ and define in the same way as above. A training example $z_i$ having \emph{positive} influence on $z_\testpred$ will \emph{decrease} the loss of the prediction when upweighted. 
% Thus, while such a training example can be considered helpful when the prediction is accurate, it is harmful when the prediction is inaccurate.
% TODO: Determine if we even need to introduce this.
% TODO: think about regression setting.

\subsection{Influence for General Loss Functions}

Under certain conditions, it can be shown that
\begin{align}
    \label{eq:dee_params}
    \frac{\dee \thetaepsi}{\dee \epsilon_i}\bigg\rvert_{\epsilon_i = 0} = - H_{\thetaopt}^{-1} \gradient{i},
\end{align}
where $H_{\thetaopt} = \frac{1}{n}\sum_{i} \nabla^{2}_{\thetaopt} \loss(z_i,\thetaopt)$ is the Hessian of the objective at $\thetaopt$ 
and $\gradient{i}  = \grad_{\thetaopt} \loss(z_i,\thetaopt)$.
This result can be obtained by applying the implicit function theorem to the first-order optimality conditions for \cref{eq:theta_star} \citep{cook1982residuals}.
In this case, the expression for the influence of $z_i$ on $z_\testpnt$ becomes 
\begin{align}
    \label{eq:hessian_IF}
    \Infl{\testpnt}{i} = \gradient{\testpnt}^T H^{-1}_{\thetaopt} \gradient{i}.
\end{align}
Again noting that $\thetaepsi\rvert_{\epsilon_i = 0} = \thetaopt$, we can form a first-order Taylor series approximation for the change in model parameters due to upweighting $z_i$,
\[
\label{eq:IF_params}
\thetaepsi -\thetaopt
    \approx \frac{\dee \thetaepsi}{\dee \epsilon_i}\bigg\rvert_{\epsilon_i = 0} \epsilon_i
    =  - H_{\thetaopt}^{-1} \gradient{i} \epsilon_i.
\]
Similarly, we can approximate the change in loss via
\[\label{eq:differenceinloss}
\nonumber
\loss(z_{\testpnt},\thetaepsi) -\loss(z_{\testpnt},\thetaopt)
    &\approx  \frac{\dee \loss(z_{\testpnt},\thetaepsi)}{\dee \epsilon_i} \epsilon_i \\
    % \nonumber
    %&= -\Infl{\testpnt}{i} \epsilon_i\\
    &= - \gradient{\testpnt}^T H^{-1}_{\thetaopt} \gradient{i} \epsilon_i.
\]

% The closed-form expression stated in \cref{eq:IF_params} can be employed to compute the influence of $z_i$ on functions of $\theta$. 
% In particular, we can compute the change in loss on some example $z_{\testpnt}$ of interest due to change in $w_i$. 
% The first order approximation and an application of the chain rule yields

for an infinitesimal change $\epsilon$ in $w_i$. 
The last equation follows from \cref{eq:IF_params}.

% For example, the effect that removing example $z_i$ will have on the loss of $z_{\testpnt}$ after another few dozen epochs of training is correlated with $\Infl{\testpnt}{i} = \frac{1}{n}\gradient{\testpnt}^T H^{-1}_{\thetaopt} \gradient{i}$.
% TODO: Why is this line breaking compilation?

%----------------------------------------------------------------------
\subsection{Influence in Maximum Likelihood}
\label{sec:connecttofisher}
%----------------------------------------------------------------------
% We have presented influence functions \NA{from the perspective of optimizing} \fTBD{DR: huh?}
% the empirical average of a generic twice continuously differentiable loss function over a set of examples.
We have presented influence functions as they have been derived for the analysis of a broad class of empirical risk minimization objectives.
% \NA{However, a few variations of the same essential concept have been examined from different perspectives (e.g., \citep{jaeckel1972infinitesimal,efron1982jackknife, efron1981nonparametric,giordano2018swiss}} \fTBD{GKD: Not sure Tamara's work is relevant here}). 
In this section, we study the special case of cross entropy loss and maximum likelihood estimation (MLE), 
which leads us to an alternative representation of IF in terms of Fisher information.
We use this formulation of IF to derive relative influence in \cref{sec:relatif}.

Let $p_{\theta}(y|x)$ be a parametric statistical model (i.e., likelihood) and let $\loss((x,y),\theta) = - \log p_{\theta}(y|x)$. Global optimization of \cref{eq:theta_star} corresponds to maximum likelihood estimation.
As shown in \citet{ting2018optimal},
in this setting the influence function satisfies
\begin{align}
    \label{eq:fisher_IF}
    \Infl{\testpnt}{i} = \gradient{\testpnt}^T F^{-1}_{\thetaopt} \gradient{i},
\end{align}
where 
\[
F_{\theta} = \frac{1}{n} \sum_{i=1}^{n} \E_{p_{\theta}} \nabla_{\theta} \log p_{\theta}(y|x_i) \nabla_{\theta} \log p_{\theta}(y|x_i)^T
\]
is the Fisher information matrix.\footnote{We do not model $p(x)$ and so we replace it with an empirical estimate using training data $S$}
Then a Taylor series approximation yields
\begin{align}
    \label{eq:fisher_IF_params}
    \thetaepsi -\thetaopt &\approx -F^{-1}_{\thetaopt} \gradient{i} \epsilon_i, \text{~~and} \\
    \label{eq:fisher_diffinloss}
    \loss(z_{\testpnt},\thetaepsi) -\loss(z_{\testpnt},\thetaopt) &\approx -\gradient{\testpnt}^T F^{-1}_{\thetaopt} \gradient{i} \epsilon_i.
\end{align}
We briefly discuss the relationship between $F_{\thetaopt}$ and $H_{\thetaopt}$ in \cref{app:connectfisher}.
For a more thorough treatment, see \citep{martens2014new}.

%----------------------------------------------------------------------
\section{Relative Influence}
\label{sec:relatif}
%----------------------------------------------------------------------
\label{sec:normIF}
For a given test input $x_\testpnt$, we would like to identify the training examples that serve to explain the target $y_\testpnt$.
The change in loss on $z_{\testpnt}=(x_\testpnt, y_\testpnt)$ due to re-weighting each $z_i$ in $S$ is described by $\{\Infl{\testpnt}{i}\}_{i=1,...,n}$.
\NA{Returning the top-k examples that maximize $\abs{\Infl{\testpnt}{i}}$ constitutes one form of explanation.
However, top influential examples selected via influence functions are often high-loss, i.e., they are either mislabelled, outliers, or otherwise atypical. 
The sets of top-k examples which maximize $\abs{\Infl{\testpnt}{i}}$ for different test inputs also often overlap.
We hypothesize that these types of examples are returned because maximizing $\abs{\Infl{\testpnt}{i}}$ puts no constraint on how re-weighting these examples affects the overall model.}

In this section we introduce relative influence, a set of measures which, using influence functions, quantify changes in loss subject to constraints on how the model may change.
We show the training examples returned under these constraints are lower loss, more specific to the test input, and that they arguably constitute a more intuitive explanation for an end user.

%----------------------------------------------------------------------
\subsection{$\theta$-Relative Influence}
%----------------------------------------------------------------------

The change in model parameters $\frac{\dee \thetaepsi }{ \dee \epsilon_i}$ appearing in \cref{def:influence} of an influence function is unconstrained. 
In contrast, we propose to directly or indirectly constrain the change in the model parameters.

We start with a direct constraint of model parameters.
In particular, we aim to identify
\begin{align}
    \label{eq:theta_argmax}
    \argmax_{i \in \{1,\ldots,n\}} \max_{\epsilon_i} & \ \abs{\loss(z_{\testpnt},\thetaepsi) - \loss(z_{\testpnt},\thetaopt)} \nonumber\\
    \text{ s.t. } & \norm{\thetaepsi -\thetaopt}^2 \leq \delta^2.
\end{align}
Which is to ask: for some small allowable change in the model parameters $\delta$, which training example $z_i$ should we re-weight to maximally affect the loss on $z_\text{test}$? 

\begin{proposition}\label{prop:theta_relative}
Assume that \cref{eq:IF_params} and  \cref{eq:differenceinloss} hold with equality. Then \cref{eq:theta_argmax} is equivalent to 
\begin{align}
    \label{eq:normIF2}
    \argmax_{i \in \{1,\ldots,n\}} \frac{\abs{\Infl{\testpnt}{i}}}{\norm{H_{\thetaopt}^{-1} \gradient{i}}}.
\end{align}
\end{proposition}
The proof is presented in \cref{app:relatif_proofs}.

\begin{definition}
The \textit{$\theta$-relative influence} of training example $z_i$ on the loss of test sample $z_{test}$ is
\begin{align*}
   \frac{\Infl{\testpnt}{i}}{\norm{H_{\thetaopt}^{-1} \gradient{i}}}.
\end{align*}
\end{definition}

% In \cref{sec:experiments} we compare how training examples with the highest $\theta$-relative influence on $z_{test}$ compare to those selected via maximizing unconstrained IF. 

%----------------------------------------------------------------------
\subsection{$\loss$-Relative Influence}
%----------------------------------------------------------------------
Here we propose an alternative way to constrain the change in the model in a maximum likelihood setting, where our model outputs $p_\theta(y|x)$.
In particular, we focus on the expected change in squared loss (log-likelihood).
In \NA{the appendix} we show that such a constraint is equivalent to constraining the change in the Kullback-Leibler (KL) divergence between the original and $z_i$ re-weighted model. %, i.e. $\KL{p_{\thetaepsi}(y|x) }{p_{\thetaopt}(y|x)}$. 

Limiting the allowable expected change in squared loss poses an indirect constraint on $\thetaepsi$.
We aim to identify
\begin{align}
    \label{eq:loss_argmax}
    \nonumber
    \argmax_{i \in \{1,..,n\}}\, \max_{\epsilon_i} \ & \abs{\loss(z_{\testpnt},\thetaepsi) - \loss(z_{\testpnt},\thetaopt)}\\
    \text{ s.t. } & \E_{p_{\thetaopt}} (\loss(z,\thetaepsi) - \loss(z,\thetaopt))^2 \leq \delta^2.
\end{align}
Which is to ask: for some small, total allowable expected squared change in loss, which training example $z_i$ should we re-weight so as to maximally affect the loss on $z_\testpnt$?

\begin{proposition}\label{prop:loss_relative}
Assume \cref{eq:fisher_IF_params} and \cref{eq:fisher_diffinloss} hold with equality. Then \cref{eq:loss_argmax} is equivalent to 
\begin{align}
    \argmax_{i \in \{1,\ldots,n\}} \frac{\abs{\Infl{\testpnt}{i}}}{\sqrt{\Infl{i}{i}}}.
\end{align}
\end{proposition}
The proof is presented in \cref{app:relatif_proofs}.

\begin{definition}
The \textit{$\loss$-relative influence} of training example $z_i$ on the loss of test sample $z_{test}$ is
\begin{align*}
   \frac{\Infl{\testpnt}{i}}{\sqrt{\Infl{i}{i}}}.
\end{align*}
\end{definition}

%----------------------------------------------------------------------
\subsection{Geometric Interpretation}
%----------------------------------------------------------------------
\label{subsec:toy}
The influence of a training example $z_i$ on the loss of a test sample $z_{\testpnt}$ can be written as
\begin{align*}
    \Infl{\testpnt}{i} &= \inner{-\gradient{\testpnt}}{\frac{\dee \thetaepsi}{\dee \epsilon_i}}\\
    &= \inner{-\gradient{\testpnt}}{ -H_{\thetaopt}^{-1}\gradient{i}},
\end{align*}
where $\langle \cdot,\cdot \rangle$ is the inner product operator. 
The inner product notation emphasizes that $\abs{\Infl{\testpnt}{i}}$ is equal to the projection length of the change in parameters vector onto the test sample's (negative) loss gradient.
A training example $z_i$ which causes a \textit{large magnitude} change in parameters (e.g., an outlier) 
is likely to have a large magnitude influence on a wide range of test samples, i.e., such a $z_i$ has a global effect. 
Conversely, a training example $z_i$ which causes a small magnitude change in parameters, will only have a high magnitude influence on test samples for which this change in parameters is \textit{directionally aligned} with $\gradient{\testpnt}$. 

\begin{figure}[!ht]
	\centering
	\includegraphics[width=0.9\linewidth]{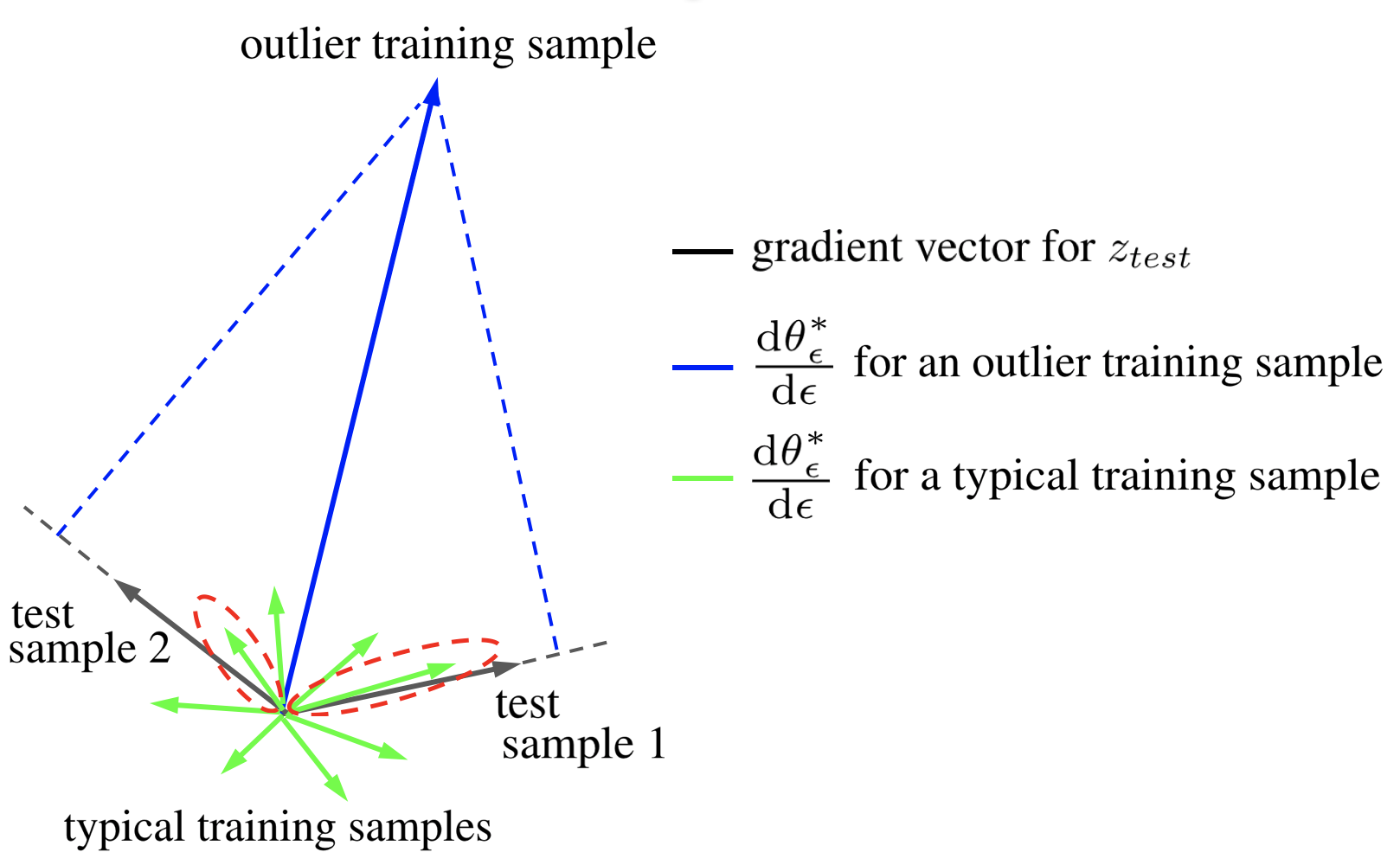}
	\caption{Geometric interpretation of \RelIF{}. 
	Using IF, the influence of $z_i$ on $z_\testpnt$ is equal to the projection length of the change in parameters vector onto the test sample's (negative) loss gradient. 
	Observe how the same outlier training example becomes the top influential example for two different test samples due to the very large magnitude effect on the model parameters.
	\RelIF{}, instead uses cosine similarity to identify the top influential examples, and thus return those in the red dotted circles.} 
	%Notice these training examples are more test-point specific.}
	\label{fig:toy_example}
\end{figure}

\RelIF{} considers the influence of a training example relative to its global effects. 
Geometrically, this switches from a paradigm of identifying relevant training examples using projections to one using cosine similarity\footnote{Recall, $\cos(a,b) = \frac{\inner{a}{b}}{\norm{a}\norm{b}}$, and $\norm{a} = \sqrt{\inner{a}{a}}$.}.
\cref{fig:toy_example} illustrates this difference.

%We can see that identifying examples using 
The equivalence between \ThetaRelIF{} and the cosine similarity between a test point gradient $\gradient{\testpnt}$ and a change in the parameters vector $\frac{\dee \thetaepsi}{\dee \epsilon_i}$ is straighforward. 
The norm of the test sample gradient $\norm{\gradient{\testpnt}}$ is independent of $i$, and so it follows that
\begin{align*}
    \argmax_i & \frac{\Infl{\testpnt}{i}}{\norm{H_{\thetaopt}^{-1} \gradient{i}}}
    = \argmax_i \frac{\inner{-\gradient{\testpnt}}{-H_{\thetaopt}^{-1}\gradient{i}}}{\norm{\gradient{\testpnt}} \norm{H_{\thetaopt}^{-1} \gradient{i}}}.
\end{align*}

Further, \LossRelIF{} can also be interpreted as cosine similarity, but not using the standard euclidean inner product.
Instead, it is cosine similarity under the inner product $\langle a, b \rangle = a^T F^{-1}_{\thetaopt} b$.
% In this space we see 
% \begin{align*}
%     \argmax_i \frac{\Infl{\testpnt}{i}}{\sqrt{\Infl{i}{i}}} = \argmax_i \frac{\inner{\gradient{\testpnt}}{\gradient{i}}}{\sqrt{\inner{\gradient{\testpnt}}{\gradient{\testpnt}}} \sqrt{\inner{\gradient{i}}{\gradient{i}}}},
% \end{align*}
% since again $\sqrt{\inner{\gradient{\testpnt}}{\gradient{\testpnt}}}$ does not depend on $i$.

\paragraph{Computational considerations}
Computing influence can be challenging, especially in larger models.
The terms in the denominator introduced by \RelIF{} pose a further computational challenge.
We outline our approach to computing relative influence in larger models in \cref{app:computation}.
It builds off the inverse Hessian vector product approximations used by \cite{koh2017understanding}, 
but makes some further approximations, 
principally motivated by K-FAC~\citep{martens2015optimizing}

%----------------------------------------------------------------------
\section{Experimental Results}
%----------------------------------------------------------------------
\label{sec:experiments}
\begin{table*}
	\centering
	\resizebox{\linewidth}{!}{
	\begin{tabular}{|l|c|c|c|c||c|c|c|c|}
	    \cline{2-9}
	    \multicolumn{1}{c|}{}&\multicolumn{4}{c||}{Logistic Regression on MNIST} & \multicolumn{4}{c|}{ConvNet on CIFAR10}\\
	    \cline{2-9}
		\multicolumn{1}{c|}{}& \multicolumn{2}{c|}{\multirow{2}{1.5cm}{Infl. Set Cardinality}} & \multicolumn{2}{c||}{\multirow{2}{1.5cm}{Infl. Set Probability}} & \multicolumn{2}{c|}{\multirow{2}{1.5cm}{Infl. Set Cardinality}} & \multicolumn{2}{c|}{\multirow{2}{1.5cm}{Infl. Set Probability}}\\
		  \multicolumn{1}{c|}{}&\multicolumn{2}{c|}{} & \multicolumn{2}{c||}{}&\multicolumn{2}{c|}{} & \multicolumn{2}{c|}{}\\
		  \hline
		 Method & top 1 & top 5 & top 1 & top 5 & top 1 & top 5 & top 1 & top 5\\
		 \hline
		 IF & 2704 & 4984 & $0.168\pm0.16$ & $0.268\pm0.21$ & 2279 & 4869 & $0.431\pm0.21$ & $0.511\pm0.21$ \\
		 \hline
		 \ThetaRelIF & 8582 & 29317 & $0.929\pm0.16$ & $0.922\pm0.17$ & 8264 & 26381 & $0.918\pm0.15$ & $0.910\pm0.16$ \\
		 \hline
		 \LossRelIF{} & 8776 & 30431 & $0.938\pm0.16$ & $0.930\pm0.17$ & 8336 & 26550 & $0.931\pm0.14$ & $0.920\pm0.15$ \\
		 \hline
	\end{tabular}}
	\caption{Comparing the set of top influential training examples recovered using IF and \RelIF{} for a logistic regression trained on MNIST and a ConvNet trained on CIFAR10 data sets. The number of test samples for each data set is $10000$. For each test sample, the top (1 or 5) positively influential training example is recovered and added to the \textit{influential set} (duplicates are removed). The cardinality of this set as well of the mean probability ($\pm$std-dev) of its elements are reported for IF and different variations of \RelIF{}. The larger cardinality of the set is an indication of recovering more specific (as opposed to globally) influential examples. Higher likelihood for the elements of the set shows more typical examples (and fewer outliers) are retrieved.\TBD{change std to SE} }
	\label{tab:infl_set_stats}
\end{table*}
We empirically assess the use of relative influence (\RelIF{}) as a technique for explaining model predictions.
We compare the examples identified by \RelIF{} to the baseline examples identified using influence functions (IF) or k-nearest neighbors (k-NN).
We consider three models and data sets: logistic regression trained on MNIST handwritten digit data set~\citep{lecun1998gradient},\footnote{The test accuracy of this model is 91.93 and the damping coefficient for hessian inversion is 0.001.}
a convolutional neural network (ConvNet) trained on CIFAR10 object recognition data set~\citep{krizhevsky2009learning},\footnote{The model architecture is 16C-32C-64C-F with max pooling after conv layers. The model test accuracy is 73.22 and the damping coefficient for hessian inversion is 0.1.}
and a long short-term memory (LSTM) network trained for classifying names based on their language of origin.\footnote{The data set is publicly available at https://download.pytorch.org/tutorial/data.zip.}\footnote{We randomly selected $15\%$ of the examples from each class for the test set. The accuracy on this test set is 71.45 and the damping coefficient for hessian inversion is 0.001 }

%+++++++++++++++++++++++++++++++++++++++++++
\subsection{Quantitative Analysis}
%+++++++++++++++++++++++++++++++++++++++++++
\label{sec:quantitative}
It is known that the training examples selected via IF are often high-loss, i.e., they are either mislabelled, outliers, or a-typical \citep{hampel1974influence, campbell1978influence,cook1982residuals,croux2000principal}.
We further demonstrate that the sets of top-k examples maximizing $\Infl{\testpnt}{i}$ for different test inputs often have a big overlap.
We quantify these properties and contrast them to the examples selected via \RelIF{}.

\paragraph{Overlap and probability}
We train a logistic regression model on MNIST, as well as a ConvNet on CIFAR10.
In each setup we identify the top-k positively influential examples for 10,000 test inputs using both IF and \RelIF{}.
For each of the selected examples, we examine the class probabilities returned by the model. Since these models are trained based on a negative-log-likelihood loss function, output probabilities are indicative of the loss.
We also examine the extent to which these examples overlap by considering the cardinality of the set of top-k examples across all 10,000 test inputs.
The maximum feasible cardinality is $k \times 10,000$ (limited by the number of training examples),
which is achieved when unique training examples are identified as explanatory for each of the test samples.
A cardinality smaller than this maximum implies that there is an overlap between the influential examples.
The results are tabulated in \cref{tab:infl_set_stats}.

We find that in all of our experiments, the positively influential points identified by \RelIF{} have little overlap.
They also have much higher probability under the model's conditional distribution, and hence have lower loss.
For example, in our ConvNet/CIFAR10 setup, the top-5 highest influential training examples identified by IF across all 10,000 test inputs is a set of only 4,869 examples with mean probability 0.511.
In contrast, when identified with \RelIF{}, the set contains on average 26,430 examples with mean probability 0.928.

\begin{table}[!t]
	\centering
  %% Including Ratios
% 	\resizebox{\linewidth}{!}{
% 	\begin{tabular}{|l|c|c|c|c|c|}
% 	    \hline
% 		Method & $\Delta\loss_{\testpnt}$ & $\norm{\Delta\theta}$ & $\sqrt{\sum(\Delta\loss_i)^2}$ & $\Delta\loss_{\testpnt}/{\norm{\Delta\theta}}$  & $\Delta\loss_{\testpnt}/\sqrt{\sum(\Delta\loss_i)^2}$ \\
% 		\hline
% 		IF &  $0.0106\pm0.0023$ & $0.039\pm0.003$ & $0.879\pm0.092$ &$0.311\pm0.167$ & $0.016\pm0.008$ \\
% 		\hline
% 		\ThetaRelIF & $0.0084\pm0.0027$ & $0.004\pm0.001$ & $0.080\pm0.033$ & $0.459\pm0.226$ & $0.026\pm0.010$ \\
% 		\hline
% 		\LossRelIF{} & $0.0086\pm0.0028$ & $0.004\pm0.002$ & $0.068\pm0.036$ & $ 0.475 \pm0.229$ & $0.027\pm0.010$ \\
% 		\hline
% 		Nearest-N & $0.0048\pm0.0025$ & $0.003\pm0.001$ & $0.044\pm0.024$ & $0.304 \pm0.182$ & $0.018\pm0.009$ \\
% 		\hline
% 	\end{tabular}
% 	}
%   %% Non-ratios
    \resizebox{\linewidth}{!}{
	\begin{tabular}{|l|c|c|c|}
	    \hline
		Method & $\Delta\loss_{\testpnt}$ & $\norm{\Delta\theta}$ & $\sqrt{\sum(\Delta\loss_i)^2}$ \\
		\hline
		IF &  $0.0106\pm0.0023$ & $0.039\pm0.003$ & $0.879\pm0.092$ \\
		\hline
		\ThetaRelIF & $0.0084\pm0.0027$ & $0.004\pm0.001$ & $0.080\pm0.033$ \\
		\hline
		\LossRelIF{} & $0.0086\pm0.0028$ & $0.004\pm0.002$ & $0.068\pm0.036$ \\
		\hline
		Nearest-N & $0.0048\pm0.0025$ & $0.003\pm0.001$ & $0.044\pm0.024$ \\
		\hline
	\end{tabular}
	}
	\caption{The change in loss at a test sample ($\Delta\loss_{\testpnt}$), compared with the norm of the change in parameters ($\norm{\Delta\theta}$) and root sum of square change in loss over the training set ($\sqrt{\sum(\Delta\loss_i)^2}$). The results come from removing the mostly positively influential training sample as determined by different methods, then retraining the model (i.e., leave-one-out retraining). The model is a logistic regression trained on MNIST. The experiment is repeated for 100 randomly selected test samples. The mean $\pm$ standard error are reported. While IF identifies the training samples having the largest influence on the test sample's loss, RelatIF finds comparably influential samples with an order of magnitude smaller global effects. A nearest neighbor baseline is included for comparison.}
	\label{tab:delta_model}
\end{table}

\paragraph{Global effects}
We are also interested in quantifying the global effect of the top influential examples selected by each method.
We train a logistic regression model on MNIST to convergence, explicitly seeding all sources of randomness.
We then pick 100 test samples uniformly at random.
For each of these test samples, we identify the most positively influential training example using IF and \RelIF{}.
We then retrain the model without that training example, maintaining the same random seed, and measure the effects.
Specifically, we consider how the removal of that training example affects the root squared error across the training set, as well as how it affects the learned model parameters. 
The results are tabulated in \cref{tab:delta_model}.

We find that the examples identified by IF have a much stronger global impact on the model than do those identified by \RelIF{}.
Using IF, the removal of the top influential examples results in a mean change in parameters of 0.039 and a root mean squared change in loss of 0.879.
Using \RelIF{}, these changes are an order of magnitude smaller.
% Note that retraining the model without the training samples identified by IF most affects the loss at the test sample. IF directly approximates the change in loss, and is thus expected to do better under this metric than \RelatIF{}, that was derived for a different purpose. However, on average, removing the samples identified by \RelatIF{} still affect the loss at z_test more than removing the nearest neighbour.

%
%+++++++++++++++++++++++++++++++++++++++++++
\subsection{Qualitative Analysis}
%+++++++++++++++++++++++++++++++++++++++++++
\label{sec:qualitative}
\NA{The purpose of our qualitative experiments is to evaluate \RelIF{} in the context of example-based explanations of model predictions.}
%Our hope is for \RelIF{} to provide insight about the \NA{reasoning process} of the model by identifying most ``relevant'' training examples (i.e., examples with most relative influence) for the generated prediction.
%These identified examples can be served as an explanation for the model prediction.
\RelIF{} is derived by explicitly minimizing the global effect on the model.
As a result, the examples selected by \RelIF{} are specific to the given test sample, and the effect of examples with large global influence is diminished. 
%To make sure the generated explanations are specific to the given test sample, \RelIF{} filters out examples with large global influence by constraining the allowable change in the model.

\begin{figure}[!h]
	\centering
	\hspace{-0.5cm}
	\includegraphics[width=1\linewidth]{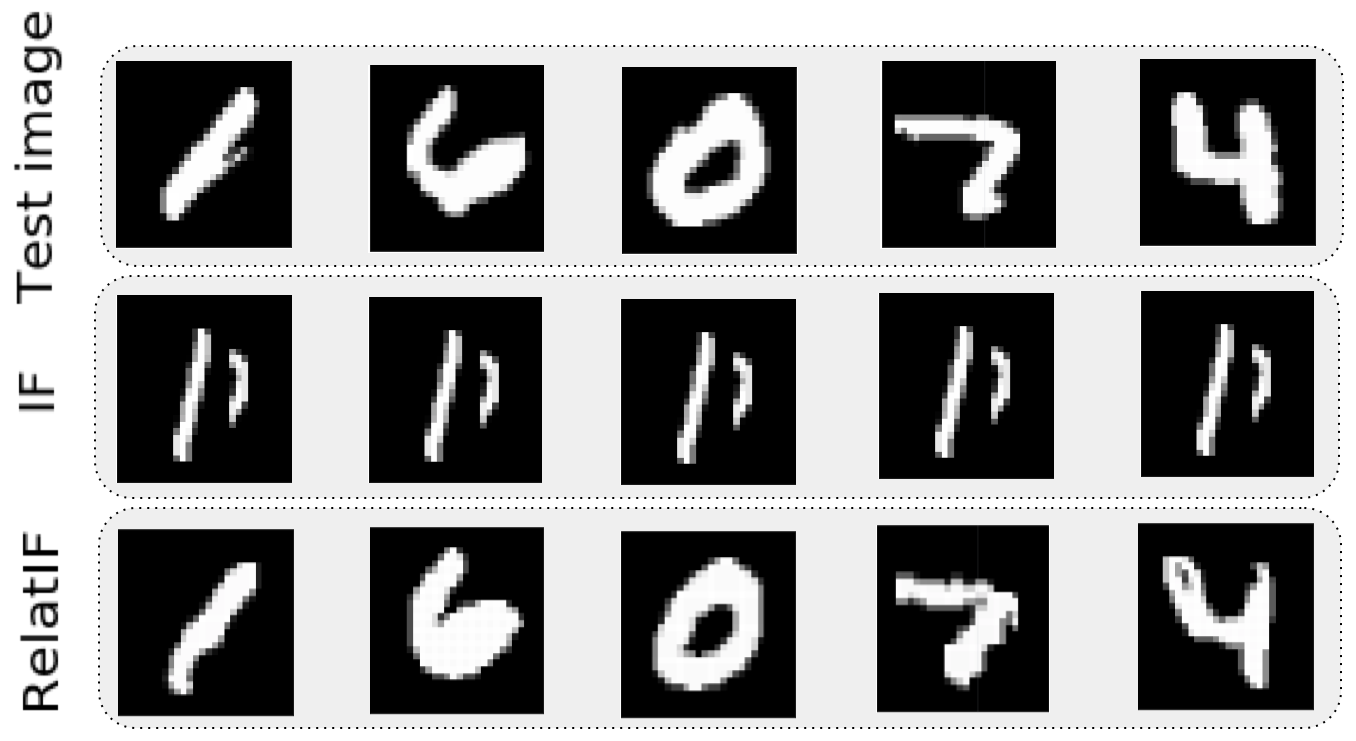}
	\caption{The most positively influential training examples suggest by IF and \RelIF{}. Test samples are in the first row. IF selects the same (high-loss) training example as the top influential example for all the test images. In contrast, \RelIF{} discovers training examples that are specific to each of the test images.}
	\label{fig:outlier}
\end{figure}

We find that examples identified by IF have significantly more global influence. % effect on the model.
An illustration of this effect for a logistic regression trained on MNIST is presented in~\cref{fig:outlier}.
As this figure shows, based on IF, a single high-loss training example is the most positively influential example for a number of different test images.
We argue that explaining the model prediction for all of these different test samples using a single high-loss (e.g., outlier) training example is not intuitive and may undermine the user's trust.
\RelIF{} addresses this issue and enables us to identify examples that are more relevant to each specific test input.\footnote{As shown in \cref{app:quant_relatif}, \ThetaRelIF{} and \RelIF{} produce similar example-based explanations and for the experiments in this section we used \ThetaRelIF{}.}

\begin{figure*}
	\centering
	\includegraphics[width=1\linewidth]{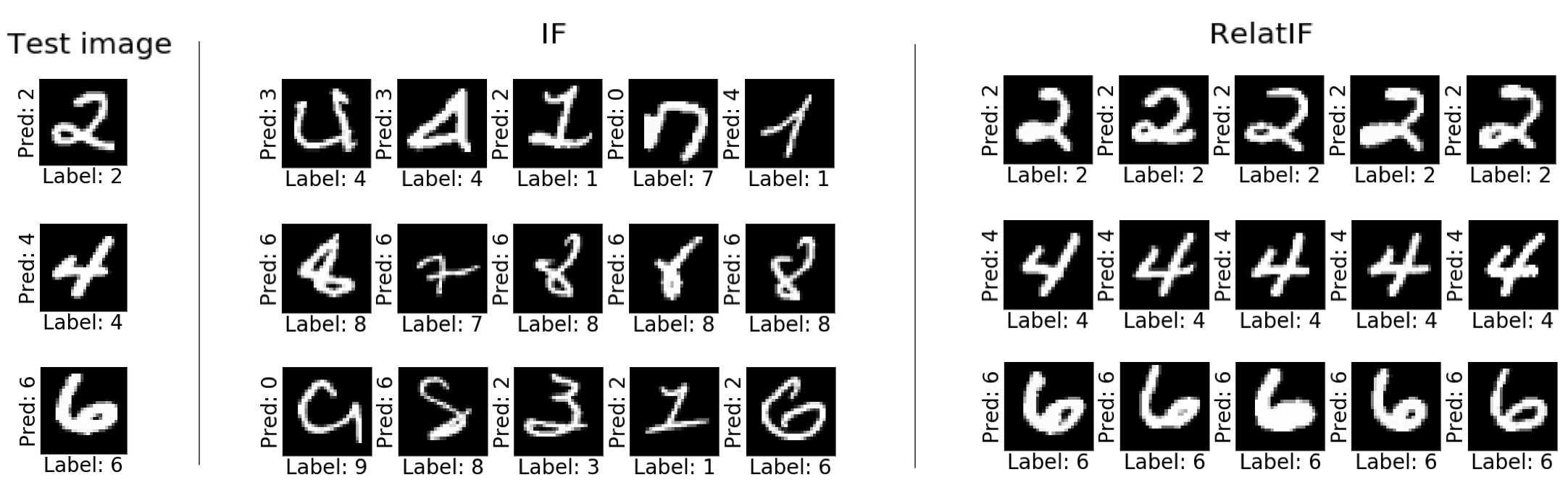}
	\caption{Comparison of the top positively influential training examples recovered by IF and \RelIF{}
	for logistic regression on MNIST. 
	Each row shows a test sample and the top five positively influential training examples for the predicted label selected by each method. The true class and the predicted label for each example are provided. Observe how most of the examples selected by IF are unusual and misclassified, despite the training set containing examples with higher visual similarity (as suggested by \RelIF{}).}
	\label{fig:exp_mnist}
\end{figure*}

We study the effectiveness of \RelIF{} for generating example-based explanations under different types of models and data sets.
\cref{fig:exp_mnist} shows the explanations produced for a logistic regression trained on MNIST.
For each test image, the top five positively influential training examples for the predicted label identified by IF and \RelIF{} are presented.
Most of the examples selected by IF are misclassified training examples
(due to their large global influence on the model parameters).
In contrast, \RelIF{} identifies visually similar examples.
We understand this to be the result of constraining global influence, thereby putting more importance on the similarity of the test/train gradients.
% We have begun to explore the extent to which, under a \RelIF{} criterion, the most influential points are also the nearest neighbors.
% Indeed, we have found that \RelIF{} does return training samples which are closer in the input space (as compared to IF).
% However, the ranking of relative influence cannot be determined solely by considering distance in the input space.
% This is a natural byproduct of prioritizing examples with maximum influence
% on the \NA{local region of interest} \fPROBLEM{DR: No! Wrong intuition.}
% in the input space (i.e., the test input).
% Note that this visual similarity is a reflective of the model \NA{reasoning process} \fPROBLEM{DR: I wouldn't go there.}
% and it is actually the type of the similarity that is captured by the model.

\begin{figure*}[!t]
	\centering
	\includegraphics[width=\linewidth]{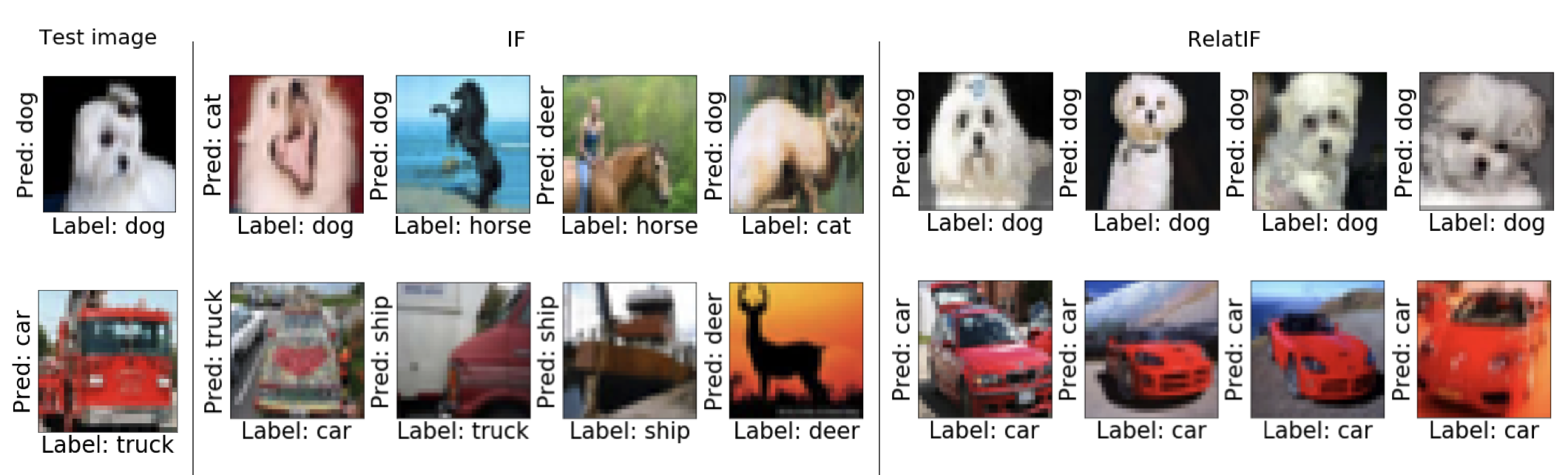}
% 	\begin{tabular}{c}
	
	   % \includegraphics[width=\linewidth]{./figures/exp_5570.png}\\
	   % \hline\\
	   % \includegraphics[width=\linewidth]{./figures/exp_7075.png}\\
% 	\end{tabular}
	\caption{Generating example-based explanations using IF and \RelIF{} for the predicted label by a ConvNet trained on CIFAR10. Each row shows a test sample and the top five positively influential training examples for the predicted label selected by each method. The true class and the predicted label for each example is marked. Compared to IF, \RelIF{} is able to retrieve training examples with higher visual similarity to the test image.}
	\label{fig:exp_cifar}
\end{figure*}

Now we make the model a bit more complex and try to explain a ConvNet trained on CIFAR10 data set.\footnote{This model has 24234 parameters and all of them are used for computing (relative) influence scores}
\cref{fig:exp_cifar} shows the generated explanation for a correctly classified (dog) and a misclassified (truck) test sample.
% For each test image in the left column, the explanation for the predicted label produced by IF and \RelIF{} is presented in the second and third column, respectively.
Interestingly, we can see that a set of visually similar dog training images are specifically helpful for predicting the label dog for the given test image. 
In the case of the red truck, 
the explanation by \RelIF{} suggests that the presence of a number of red cars (with similar shade of color and shape) specifically influences the model to classify it as a car. 
For more examples of CIFAR10 explanations see \cref{app:exp_examples}.

A similar experiment is conducted for a character-level LSTM network trained for classifying names based on their language of origin.
The data set consists of $20,000$ surnames from $18$ different languages.
Due to the large number of model parameters, similar to~\citet{koh2017understanding}, in this experiment we used only the last layer parameters for computing influence scores.
\cref{tab:exp_text} shows the generated explanation for classifying surname ``Vasyukov'' as Russian.
The examples identified by IF are all Russian names that are misclassified as Japanese.
In contrast, all of the examples identified by \RelIF{} are Russian names that end with ``kov'', similar to the test input. 

The explanations provided for these models enable the end user to assess the reliability of the generated prediction.
The user can confirm, reject or modify the model decision by inspecting the extracted evidence from the training set. 
This is especially useful when the model is not confident in its prediction.
For an example of this application see \cref{app:hitl}.

\begin{table}[!t]
	\centering
	\resizebox{1\linewidth}{!}{
	\begin{tabular}{|c||c|c|}
	    \cline{2-3}
	    \multicolumn{1}{l||}{}& IF & \RelIF{}\\
	    \hline
	    \multirow{6}{1.7cm}{\begin{tabular}{@{}c@{}} Test Input: \\ \textbf{Vasyukov} \\ (Russian) \end{tabular}}
		& Tokovoi (Japanese) & Gasyukov (Russian) \\
		\cline{2-3}
		& Nasikan (Japanese) & Tsayukov (Russian)\\
		\cline{2-3}
		& Bakai (Japanese) & Tzarakov (Russian)\\
		\cline{2-3}
		& Mihailutsa (Japanese) & Tzayukov (Russian)\\
		\cline{2-3}
		& Jugai (Japanese) & Haryukov (Russian)\\
		\cline{2-3}
		& Mikhailutsa (Japanese) & Tsarakov (Russian)\\
		\hline
	\end{tabular}
	}
	\caption{Generating Explanation for text classification. The model is a character-level RNN trained for classifying names based on their language of origin. The most positively influential training examples selected based on IF and \RelIF{} are listed. The language in parentheses shows the predicted label. The true label for all examples shown is Russian. Observe how the examples identified by IF are all misclassified as Japanese, while the examples identified by \RelIF{} end with ``kov'', which is similar to the test input. }
	\label{tab:exp_text}
\end{table}

%+++++++++++++++++++++++++++++++++++++++++++
\subsection{Comparison to Nearest-Neighbors}
%+++++++++++++++++++++++++++++++++++++++++++
Here we compare a test sample's k-nearest neighbors (k-NNs) to the samples returned using RelatIF.
For a quantitative comparison we use the same MNIST and CIFAR10 experimental setups discussed in \cref{sec:quantitative}. 
We consider the overlap between the top-k samples identified by RelatIF, and the k-NNs ($l_2$ distance) for k in $\{1,5,10,20\}$. 
In the logistic regression model on MNIST we find the overlap ranges from 26-35\%, increasing with k. 
A substantial overlap, but still indicative of a fundamentally different set of points. 
In the CNN on CIFAR10 we find much less overlap, only 2.6-5.6\%, decreasing with k. 
By comparison, the overlap between the top-k samples identified by IF and the k-NNs is 1.3-2.7\% in the logistic regression model, 
and 0.67-1.24\% in the CNN. 
This indicates that the points returned by RelatIF are closer in $l_2$ distance to the test sample than those returned by IF.

A qualitative comparison finds a significant difference between the k-NNs and the samples returned by RelatIF (see \cref{app:relatif_vs_knn}). 
These examples are abundant and easily found for our CNN on CIFAR10.
We have included some of these examples in the appendix.

%----------------------------------------------------------------------
\section{Related work}
%----------------------------------------------------------------------
Influence functions (IF) were originally proposed by \citet{hampel1974influence}. 
Methods based on IF are identical to the infinitesimal jackknife,
which was introduced earlier in an independent work on robust statistics by \citet{jaeckel1972infinitesimal}.
%in the pursuit of robust statistics.
The connection between IF and infinitesimal jackknife was recognized by 
\citep{efron1982jackknife}.
% They were studied extensively by \citet{cook1982residuals}.

\citet{koh2017understanding} use IF to identify influential training examples in a modern machine learning context. 
They further demonstrate that influence functions can be used to identify mislabelled data and generate adversarial training examples\citep{koh2018stronger}.
More recently, \citet{koh2019accuracy} use IF to approximate the effects of re-weighting groups of training examples.

A range of approaches have been introduced to identify training examples for interpreting predictions. 
One such approach was recently proposed by \citet{khanna2019interpreting}. 
The authors use Fisher kernels to identify points that are ``most responsible'' for a given set of predictions. 
They demonstrate that in the case of a negative log-likelihood loss function and with a specific choice of hyper-parameters, their approach coincides with IF.
% \citet{efron1981nonparametric} show that the method of infinitesimal jackknife and IF are mathematically equaivalent \NA{I haven't read the jackknife papers -- is this true?}.
% More recently, \citet{giordano2018swiss} provide error bounds for the first order approximation of infinitesimal jackknife.
Another recent contribution is due to \citet{ghorbani2019data}. The authors propose a new form of data valuation which identifies important examples or subsets of the data using Shapley values.

Other approaches to example-based explanation include the extraction of prototypical data, as well as generating counter-factual examples.
\citet{bien2011prototype} propose a method to select prototypes for interpretable classification.
\citet{kim2016examples} introduce a method for explaining the data set by identifying its prototypes and "criticisms" and argue that a-typical data must also be extracted.
% \citet{ting2018optimal} uses IF as an importance measure in sub-sampling.
\citet{chang2019explaining} introduce a novel method for generating counter-factual images to be used as an explanation. 

Finally, the effect of example-based explanations on user trust have been studied in the human-computer interaction community.
One of the most recent study is presented in \citep{zhou2019effects}.
For each test sample, \citeauthor{zhou2019effects} select examples maximizing influence.
They provide these selected training examples along with model prediction and study the effect on the users' trust in the predictive model.
A similar set of experiments can be found in \citep{cai2019effects}. 
In this study, the effect on users' trust is compared for different types of ``explanatory'' examples.
%----------------------------------------------------------------------
\section{Discussion}
%----------------------------------------------------------------------
In this work we introduce relative influence (\RelIF{}) and show that it identifies more intuitive examples than those identified by influence functions (IF). 
In particular, we demonstrate that, unlike examples identified by IF,
which tend to be atypical or misclassified, and otherwise unrelated to the test sample,
examples identified by \RelIF{} seem to be more specific to test sample. 
As a result, we believe that \RelIF{} may be superior at identifying explanatory examples that a user may find intuitive and acceptable.
The desiderata of example-based explanations are case specific,
and therefore one cannot make an absolute comparison of IF versus \RelIF{}.
A data scientist fitting a model to a data set containing spurious samples may very well wish to see the examples identified by \textit{both} IF and \RelIF{}.
We believe a large user study, which is beyond the scope of this paper, 
may shed light on the relative strengths and weaknesses of \RelIF{}, and lead to further insight.

% \TBD{EB: Difference with nearest neighbor}

\subsubsection*{Acknowledgements}
We thank the reviewers for their feedback and suggestions. We are grateful to all the input we received from our colleagues and advisors at Element AI. 

\bibliography{refs}
\bibliographystyle{refs}

%\end{document}

\clearpage

%----------------------------------------------------------------------
\appendix
%----------------------------------------------------------------------
%----------------------------------------------------------------------
\section{Connection between the Fisher and Hessian}
\label{app:connectfisher}
%----------------------------------------------------------------------
The Fisher information matrix of a conditional distribution parameterized by $\theta$, $p_{\theta}(y|x_i)$, where we do not have an explicit representation for $p(x)$, is
\begin{align*}
F_{\theta} 
    &= \frac{1}{n} \sum_{i=1}^{n} \E_{p_{\theta}} \nabla_{\theta} \log p_{\theta}(y|x_i) \nabla_{\theta} \log p_{\theta}(y|x_i)^T \\
    &= \frac{1}{n} \sum_{i=1}^{n}\Big[ \underbrace{\E_{p_{\theta}} \frac{\nabla_{\theta}^2 p_{\theta}(y|x_i)}{p_{\theta}(y|x_i)}}_{=0}  - \E_{p_{\theta}} \nabla_{\theta}^2 \log p_{\theta}(y|x_i) \Big] \\
    &= -\frac{1}{n} \sum_{i=1}^{n}\E_{p_{\theta}} \nabla_{\theta}^2 \log p_{\theta}(y|x_i).
\end{align*}
The second equality can be obtained by carrying out the differentiation $\nabla_{\theta}^2 \log p_{\theta}(y|x)$.
The third equality critically relies on the expectation being taken with respect to the model's distribution, as well as being able to switch the order of the expectation and differentiation. 

When our model has learned a distribution $p_{\theta}(y|x)$ close to the "true" distribution $p(y|x)$,
we may approximate $F_{\theta}$ by replacing $\E_{p_\theta}$ with a Monte Carlo estimate based on the target values, $y_i$, in the training set.
This gives us
\begin{align*}
F_{\theta} \approx - \frac{1}{n} \sum_{i=1}^{n} \nabla_{\theta}^2 \log p_{\theta}(y_i|x_i) = H_{\theta}.
\end{align*}
The final equality with $H_{\theta}$ (as defined in \cref{eq:dee_params}) holds when $\loss(z, \theta) = -\log p_{\theta}(y_i|x_i)$.
Therefore, our two expressions for influence, \cref{eq:hessian_IF} and \cref{eq:fisher_IF}, approximately agree when the loss is the negative log likelihood, and $y$ given $x$ is actually distributed as $p_{\thetaopt}(y|x)$.
In practice, we have found that the Fisher and Hessian return similar examples when used to compute $\Infl{\testpnt}{i}$.
%----------------------------------------------------------------------
\section{Proofs of \RelIF{} }
\label{app:relatif_proofs}
%----------------------------------------------------------------------
Recall \textbf{\cref{prop:theta_relative}}:
\textit{Assume that \cref{eq:IF_params} and  \cref{eq:differenceinloss} hold with equality. Then \cref{eq:theta_argmax} is equivalent to}
\begin{align}
    \label{eq:normIF2}
    \argmax_{i \in \{1,\ldots,n\}} \frac{\abs{\Infl{\testpnt}{i}}}{\norm{H_{\thetaopt}^{-1} \gradient{i}}}.
\end{align}

\begin{proof}
The $\argmax$ in \cref{eq:theta_argmax} is just a search over the $n$ examples in our training set. 
Therefore the solution to \cref{eq:theta_argmax} amounts to solving $\max \abs{\loss(z_{\testpnt},\thetaepsi) - \loss(z_{\testpnt},\thetaopt)}$ subject to the constraints for each $i$, then choosing the $z_i$ for which this quantity is largest. 
We assume \cref{eq:differenceinloss} holds with equality, and so it follows
\begin{align*}
 \abs{\loss(z_{\testpnt},\thetaepsi) - \loss(z_{\testpnt},\thetaopt)} = \abs{\Infl{\testpnt}{i} \epsilon_i}.
\end{align*}
Note, that $\Infl{\testpnt}{i}\epsilon_i$ is a linear function in $\epsilon_i$. 
The maximum of its absolute value will therefore be on one of the endpoints imposed by the constraint on $\epsilon_i$.

We assume that approximation in \cref{eq:IF_params} is exact, yielding $\thetaepsi -\thetaopt = - H_{\thetaopt}^{-1} \gradient{i} \epsilon_i$. 
Using this equality to describe the change in parameters, the constraint in \cref{eq:theta_argmax} can be written as
\begin{align*}
    \norm{H_{\thetaopt}^{-1} \gradient{i}\epsilon_i}^2 \leq \delta^2 \implies \abs{\epsilon_i} \leq \frac{\abs{\delta}}{\norm{H_{\thetaopt}^{-1} \gradient{i}}}.
\end{align*}
We now have an explicit constraint on $\epsilon_i$.
Plugging either endpoint of this interval into $\abs{\Infl{\testpnt}{i}\epsilon_i}$ yields the same value.
Substituting that value into the outer $\argmax$ we get,
\begin{align*}
    \argmax_{i \in \{1,\ldots,n\}} \frac{\abs{\Infl{\testpnt}{i}}\abs{\delta}}{\norm{H_{\thetaopt}^{-1} \gradient{i}}}.
\end{align*}
Since $\delta$ does not depend on $i$, it can be dropped from the $\argmax$, and the result follows.
\end{proof}

Recall \textbf{\cref{prop:loss_relative}}:
Assume \cref{eq:fisher_IF_params} and \cref{eq:fisher_diffinloss} hold with equality. Then \cref{eq:loss_argmax} is equivalent to 
\begin{align}
    \argmax_{i \in \{1,\ldots,n\}} \frac{\abs{\Infl{\testpnt}{i}}}{\sqrt{\Infl{i}{i}}}.
\end{align}

\begin{proof}
The proof follows the proof of \cref{prop:theta_relative}.
We need only consider the new constraint.
Assuming \cref{eq:fisher_diffinloss} holds with equality, we have
\begin{align*}
     \loss(z_j,\thetaepsi) - \loss(z_j,\thetaopt)
        = \Infl{j}{i} \epsilon_i
        = - \gradient{j}^T F^{-1}_{\thetaopt} \gradient{i}\epsilon_i.
\end{align*}
We introduce the notation $\gradfunc{z} = \nabla_{\theta} \loss(z,\thetaopt)$ to signify the gradient of the loss of a point $z=(x,y)$ sampled from the model's distribution, $p_\thetaopt$.
We substitute into the constraint, which yields
\begin{align*}
    \E_{p_{\thetaopt}} (- \gradfunc{z}^T F^{-1}_{\thetaopt} \gradient{i}\epsilon_i)^2 \leq \delta^2.
\end{align*}
Expanding and rearranging the left hand side, yields
\begin{align*}
        \E_{p_{\thetaopt}} (\gradfunc{z}^T & F^{-1}_{\thetaopt} \gradient{i}\epsilon_i)^T(\gradfunc{z}^T F^{-1}_{\thetaopt} \gradient{i}\epsilon_i) \\
        &=  \gradient{i}^T F^{-1}_{\thetaopt} \E_{p_{\thetaopt}} \big[\gradfunc{z} \gradfunc{z}^T \big] F^{-1}_{\thetaopt}\gradient{i} \epsilon_i^2,
\end{align*}
where we have moved the constant terms outside of the expectation.
Because our loss functions here is negative log-likelihood, $\E_{p_{\thetaopt}} [\gradfunc{z} \gradfunc{z}^T ]$ is the is the definition of the Fisher information matrix, $F_\thetaopt$. 
\NA{It agrees with the presentation in \cref{sec:connecttofisher} when we replace the expectation over $z=(x,y)$ with $\sum_j E_{p_{\thetaopt}(y|x_j)}$}.
Thus the constraint can be reduced to
\begin{align*}
    \gradient{i}^T F^{-1}_{\thetaopt} \gradient{i} \epsilon_i^2 
        =\Infl{i}{i} \epsilon_i^2 \leq \delta^2.
\end{align*}
% \begin{align*}
%     \E_{p_{\thetaopt}} (\loss(z,\thetaepsi) &- \loss(z,\thetaopt))^2 \\
%         &= \E_{p_{\thetaopt}} (- \gradfunc{z}^T F^{-1}_{\thetaopt} \gradient{i}\epsilon_i)^2 \\
%         &= \E_{p_{\thetaopt}} (\gradfunc{z}^T F^{-1}_{\thetaopt} \gradient{i}\epsilon_i)^T(\gradfunc{z}^T F^{-1}_{\thetaopt} \gradient{i}\epsilon_i)\\
%         &=  \gradient{i}^T F^{-1}_{\thetaopt} \E_{p_{\thetaopt}} \big[\gradfunc{z} \gradfunc{z}^T \big] F^{-1}_{\thetaopt}\gradient{i} \epsilon_i^2 \\
%         &= \gradient{i}^T F^{-1}_{\thetaopt} \sum_j E_{p_{\thetaopt}(y|x_j)} \big[\gradient{j} \gradient{j}^T \big] F^{-1}_{\thetaopt}\gradient{i} \epsilon_i^2 \\
%         &=  \gradient{i}^T F^{-1}_{\thetaopt} F_{\thetaopt} F^{-1}_{\thetaopt}\gradient{i} \epsilon_i^2 \\
%         &=  \gradient{i}^T F^{-1}_{\thetaopt} \gradient{i} \epsilon_i^2 \\
%         &=  \Infl{i}{i} \epsilon_i^2 \leq \delta^2.
% \end{align*}
The Fisher, $F_{\thetaopt}$, is positive definite, therefore $\Infl{i}{i}$ is positive, and the constraint is equivalent to 
$\abs{\epsilon_i} \leq \frac{\abs{\delta}}{\sqrt{\Infl{i}{i}}}$.
The rest of the proof follows \cref{prop:theta_relative}.
\end{proof}

% %----------------------------------------------------------------------
% \section{Geometric interpretation of dividing by self-influence}
% \label{app:selfIF}
% %----------------------------------------------------------------------

% Since $H_{\thetaopt}^{-1}$ is positive-definite, influence functions formulation can be written as
% \begin{align*}
%   \Infl{\testpnt}{i} &= - \langle \gradient{\testpnt} , \gradient{i} \rangle_{\Hspace}
% \end{align*}
% where $\langle u , v \rangle_{\Hspace}$ is defined as $u^T H_{\thetaopt}^{-1} v$. Therefore, influence of $z_i$ on $z_\testpnt$ is the inner product between their gradients in the Hessian space. The cosine similarity between these gradient vectors in the Hessian space becomes
% \begin{align*}
%   similarity &= \frac{-\langle \gradient{\testpnt} ,\gradient{i} \rangle_{\Hspace}}
%                               {\sqrt{\langle\gradient{\testpnt} ,\gradient{\testpnt}\rangle_{\Hspace}} 
%                                 \sqrt{\langle \gradient{i} , \gradient{i} \rangle_{\Hspace}} }  \\
%   &= \frac{\Infl{\testpnt}{i}}
%                 {\sqrt{\Infl{\testpnt}{\testpnt}}
%                  \sqrt{\Infl{i}{i}} }.
% \end{align*}
% Therefore, \cref{eq:normIF1} normalizes influence functions to retrieve training examples with high cosine similarity between $\gradient{i}$ and $\gradient{\testpnt}$ in the Hessian space.

%----------------------------------------------------------------------
\section{Computational Considerations}
\label{app:computation}
\begin{figure*}[t!]
    \centering
    \begin{tabular}{c c c}
         \includegraphics[width=0.33\linewidth]{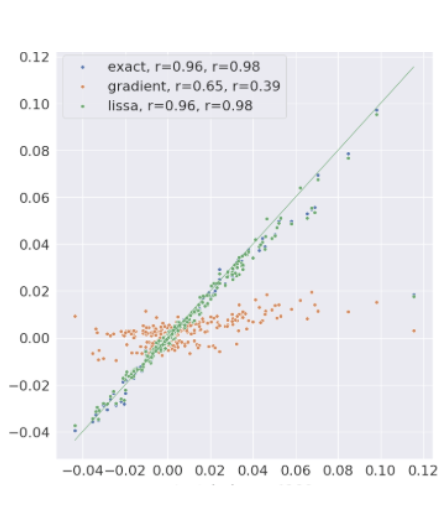}&
         \includegraphics[width=0.33\linewidth]{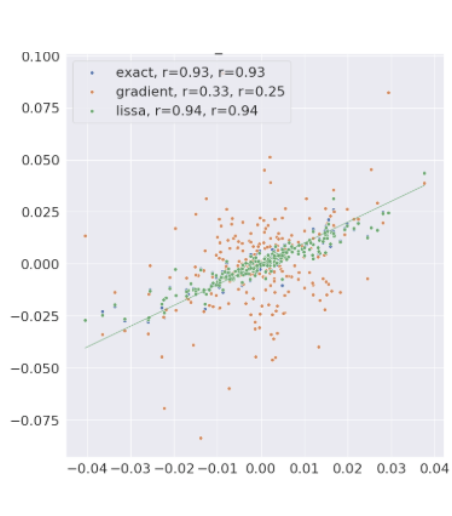}&
         \includegraphics[width=0.33\linewidth]{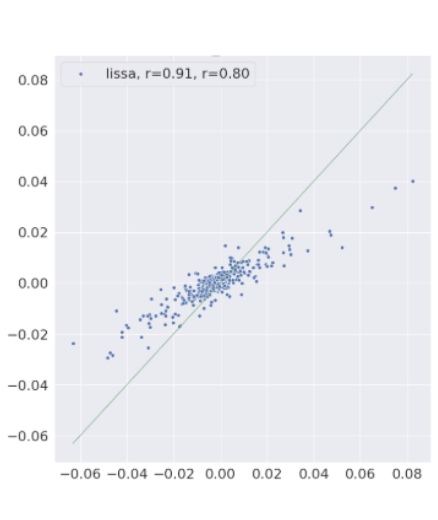}
         \\
        (a) Logistic Regression on MNIST & (b) ConvNet on MNIST & (c) AlexeNet on small ImageNet
    \end{tabular}
    \caption{Evaluating the accuracy of the estimated error for leave-one-out retraining by influence functions with different hessian approximations. in the figure, ``exact'' refers to exact Hessian, ``gradient'' referes to using identity matrix for Hessian and ``lissa'' refers to using the LiSSA method for approximating inverse Hessian vector products. In the case of AlexNet, the results are only reported for LiSSA due computational/memory complexity. For all of these models a small damping coefficient has been added to the diagonal of the Hessian to make it invertible. As this figure shows, in the case of LiSSA and exact Hessian, the correlation between influence function estimation for change of loss and the actual leave-one-out retraining loss is high. }
    \label{fig:if_eval}
\end{figure*}

%----------------------------------------------------------------------
\paragraph{Inverting the Hessian}
The Hessian of the total loss, $H_{\thetaopt}$, is positive definite if $\thetaopt$ is the local minimum of the objective function. 
If the requirements for positive definiteness of $H_\theta$ are not met, for example because the optimization procedure has not fully converged, we can add a damping term to the Hessian to make it invertible
(i.e., $\tilde{H}_\theta = {H}_\theta + \lambda I$ where $I$ is the identity matrix). 
Adding the damping term is equivalent to $L_2$ regularization, and allows us to form a convex quadratic approximation to the loss function. 
~\citet{koh2017understanding} demonstrate that influence functions computed with a damping term still give meaningful results in practice.
Empirically we have also found this to be the case (see~\cref{fig:if_eval}).
\label{app:if_eval}
\paragraph{IF and \RelIF{} in large models}
The Hessian of the total loss, $H_{\thetaopt}$, is a $P$ by $P$ matrix, where $P$ is the number of model parameters.
In even just mid-sized models, it becomes near impossible to explicitly form $H_{\thetaopt}$ in memory. 
\citet{koh2017understanding} propose using LiSSA~\citep{agarwal2017second} to estimate the inverse Hessian vector products needed to compute influence. 
We have found that this works well in practice after some tuning.
However, because the denominator in \RelIF{} is a function of the training example $z_i$ we cannot use the same $s_{\testpnt}$ trick suggested by \citet{koh2017understanding}. 
This has led us to explore using a number of block diagonal approximations to the Hessian, principally motivated by K-FAC~\citep{martens2015optimizing}.
Specifically, we use these approximations to pre-compute the denominator in \RelIF{} ($\sqrt{\Infl{i}{i}}$ or $\norm{H^{-1}_{\thetaopt}\gradient{i}}$) for every training example $z_i$.
We are then able to use $s_{\testpnt}$. 

%----------------------------------------------------------------------
\section{Comparison to support vectors}
\label{app:svms}
%----------------------------------------------------------------------

Consider a soft-SVM optimization problem, with a dimensionality of the feature space lower than the number of training points. 
Via working with a dual formulation of the objective, one can uncover that the optimal weights $w^{\star}$ lie in the linear span of some training samples, called support vectors (see, e.g., \citep{shalev2014understanding}, Ch.15).

Let $\psi(\cdot)$ map inputs $x$ to a feature space.
Define a kernel function $K(x_i,x_j) = \inner{\psi(x_i)}{\psi(x_j)}$ for all inputs $x_i,x_j$. Let $w^*$ be an output of a solution to a soft-margin SVM.

The support vectors are the training samples indexed by $i$, such that $y_i \inner{w^*}{\psi (x_i)} \leq 1$, i.e., support vectors are the training samples that are inside the margin or misclassified. 
By computing the gradients of a soft-SVM objective, one can see that IF induces a ranking of training samples according to 
$y_i y_{test} K(x_i, x_{test})$, where $\{x_i\}_i$ are support vectors (IF equals to zero for training samples that are not support vectors).  
In other words, the highest influence samples are support vectors that are most similar to the test point in the feature space. 
Similarly, repeating the same calculation for the loss-relative influence, one would rank the support vectors based on $y_i y_{test} K(x_i, x_{test}) / \sqrt{K(x_i,x_i)}$. 

In summary, IF gives a way to distinguish which of the training samples \emph{among all support vectors} are the ones that are most similar to the test sample in the feature space. 
RelatIF normalizes this similarity in the feature space by the norm of the training sample in the feature space, 
which is equivalent to IF evaluated on training samples that are unit vectors in the feature space.

%----------------------------------------------------------------------
\section{Application of example-based explanations}
\label{app:hitl}
\begin{figure*}[h]
	\centering
	\includegraphics[width=\linewidth]{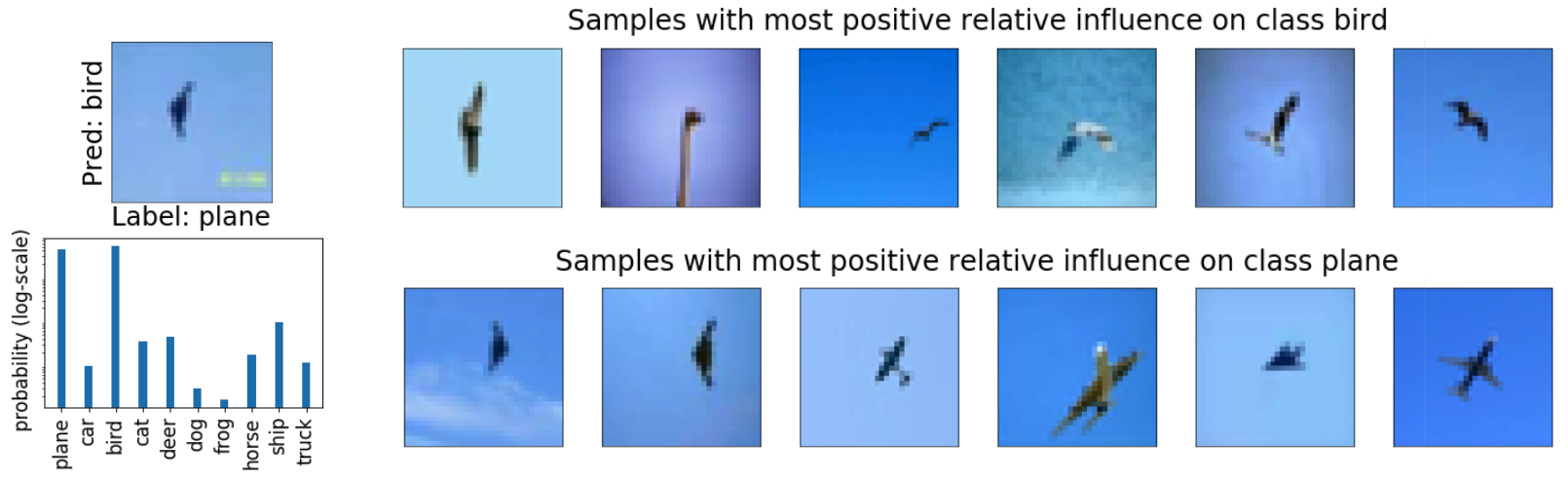}
	\caption{Using \RelIF{} to identify training examples with the top positive relevant influence on the the predicted (bird) and true label (plane).}
	\label{fig:hitl}
\end{figure*}

%----------------------------------------------------------------------
Consider the example shown in \cref{fig:hitl}.
The input image in the top left corner is classified as a bird, however, 
the predicted probability for class plane is also very high. 
Using \RelIF{} we can identify the top relevant training examples to each label (i.e., bird and plane) and assess the model's decision based on them.
Comparing the similarity between the test image and relevant training examples to each class,
suggests that this image should be classified as a plane. Also, these explanations could be used as a guide to collect more training examples for improving the model performance (e.g., what kind of training examples are useful for correcting a specific misclassification).
%----------------------------------------------------------------------
\section{Qualitative comparison of \ThetaRelIF{} and \LossRelIF{}}
\begin{figure*}[t!]
	\centering
	\includegraphics[width=1\linewidth]{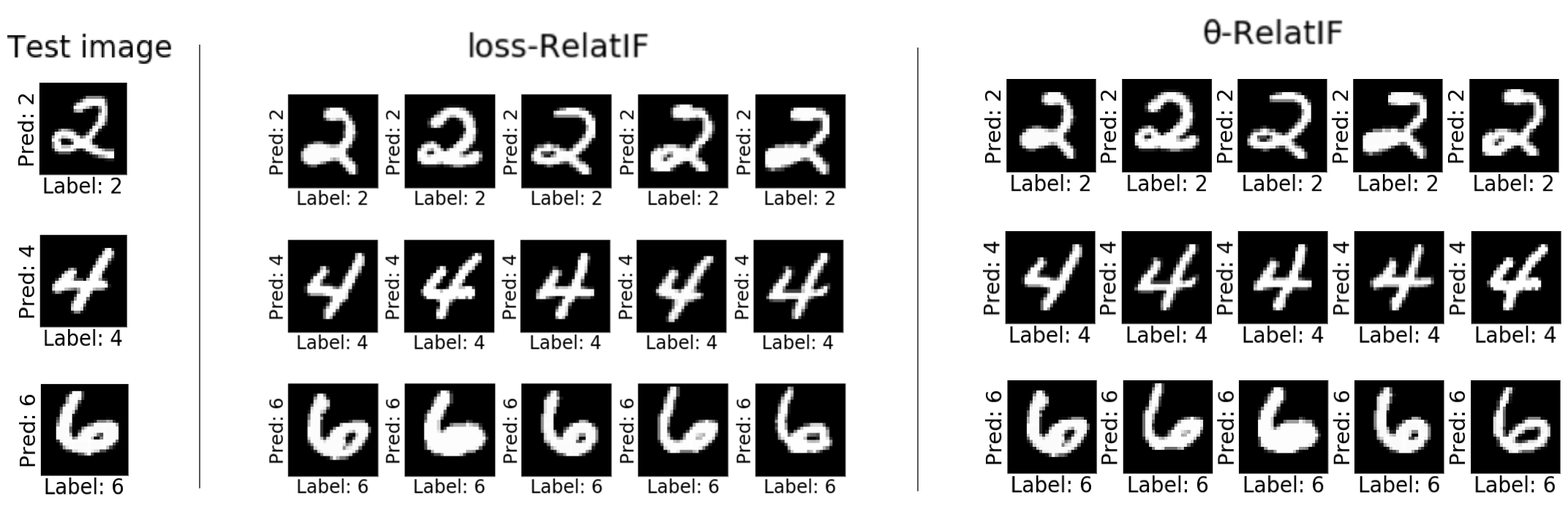}
	\caption{Comparison of the top positively influential training examples recovered by \LossRelIF{} and \ThetaRelIF{}. The classifier is a logistic regression trained on MNIST. 
	Each row shows a test sample and the top five positively influential training examples for the predicted label selected by each method. The true class and the predicted label for each example is marked. The example returned by both method are visually similar.}
	\label{fig:quant_relatif}
\end{figure*}

%----------------------------------------------------------------------
\label{app:quant_relatif}
\cref{fig:quant_relatif} offers a comparison between the top positively influential training examples recovered by \LossRelIF{} and \ThetaRelIF{}. As this figure shows, the examples returned by both method are visually similar.
%%
%----------------------------------------------------------------------
\section{Qualitative comparison with k-nearest neighbors}
\label{app:relatif_vs_knn}
\begin{figure*}[h]
\centering
\begin{tabular}{c}
     \includegraphics[width=0.9\linewidth]{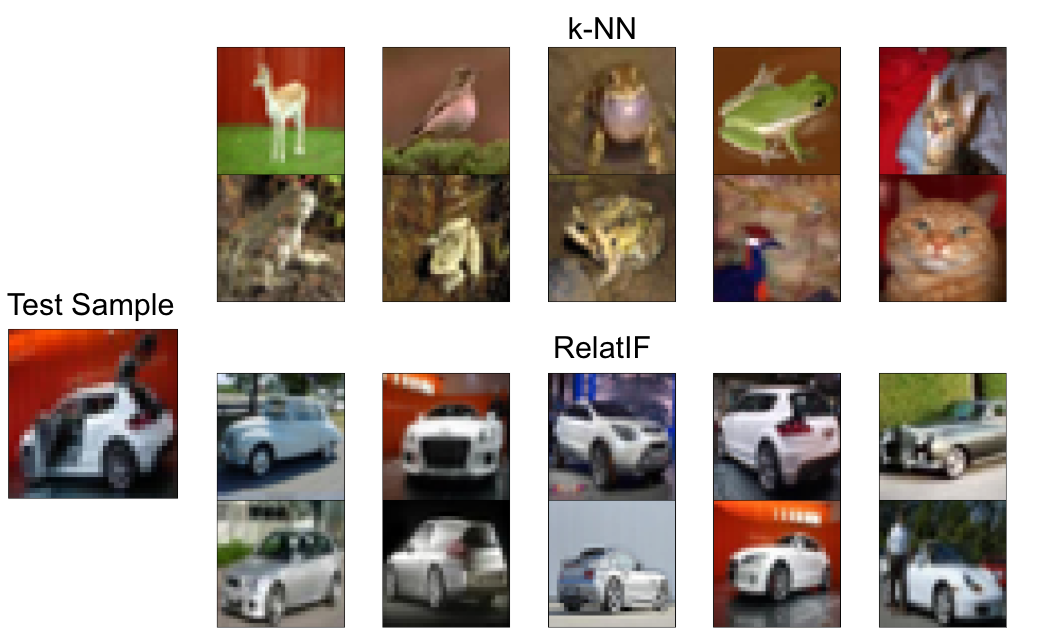}\\
     \\
     \hline
     \includegraphics[width=0.9\linewidth]{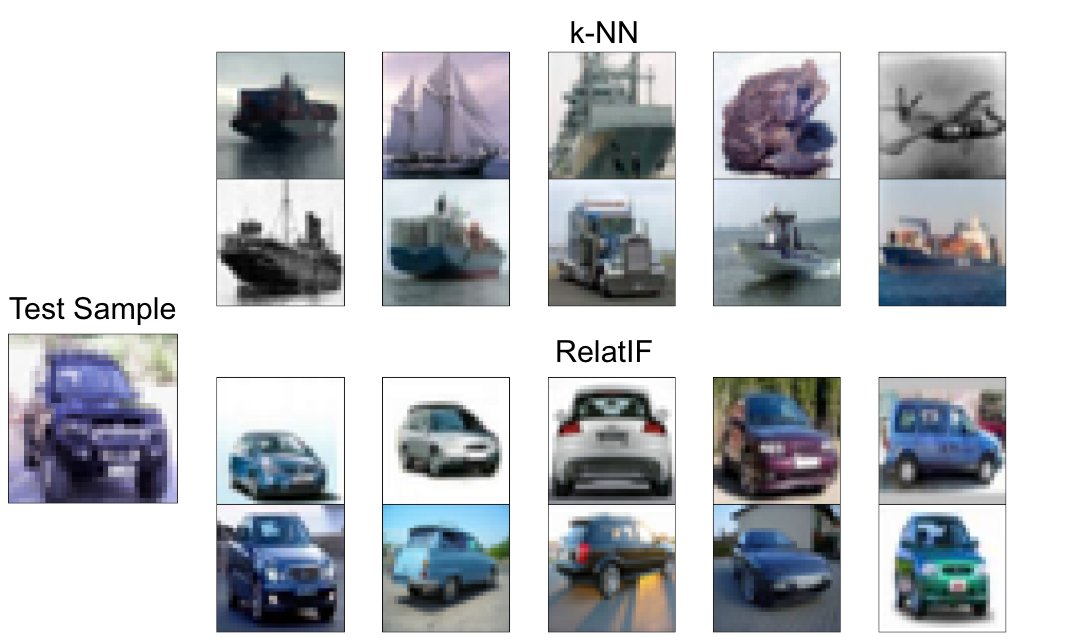}\\
\end{tabular}
\caption{Comparing the k-nearest neighbors (k-NN) to the most (positive) relatively influential samples (RelatIF) for a convolutional neural network trained on CIFAR10. The two test samples were correctly classified by the model. The k-nearest neighbors are similar in pixel composition, but are semantically different from the test sample. They provide no evidence to explain why the model has correctly classified the test sample. Conversely, the samples returned by RelatIF are of the matching class. They suggest the model has been exposed to relevant training data, and has learned something useful.}
\label{fig:relatif_vs_knn}
\end{figure*}

%----------------------------------------------------------------------
\cref{fig:relatif_vs_knn} offers a qualitative comparison between using  \RelIF{} and k-nearest neighbors for explaining the prediction of a ConvNet trained on CIFAR10 data set.
%----------------------------------------------------------------------
\section{More explanation examples}
\begin{figure*}[h]
	\centering
	\begin{tabular}{c}
	    \includegraphics[width=\linewidth]{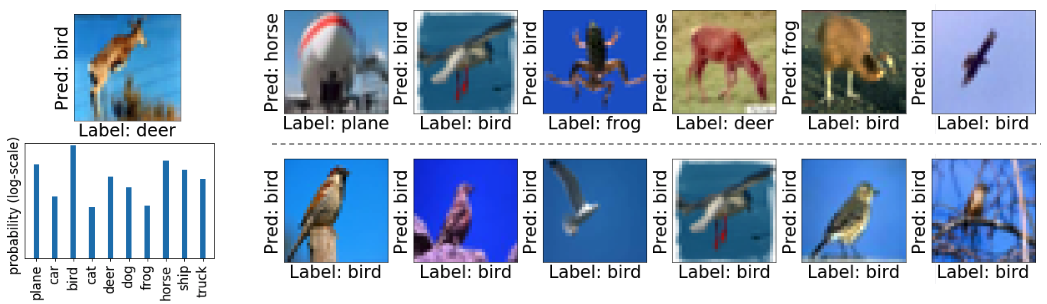}\\
	    \hline\\
	    \includegraphics[width=\linewidth]{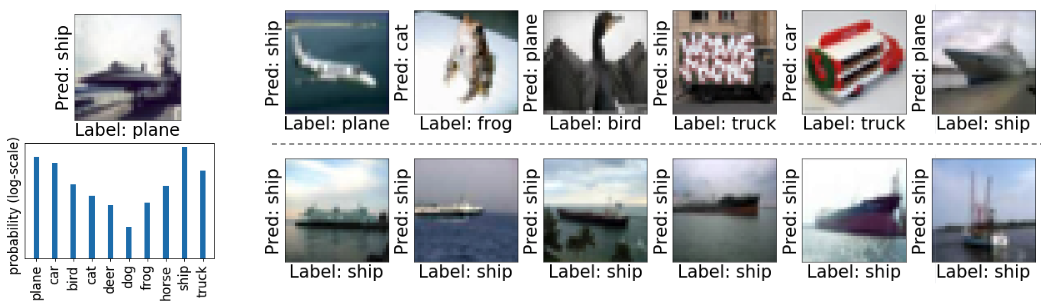}\\
	    \hline\\
	    \includegraphics[width=\linewidth]{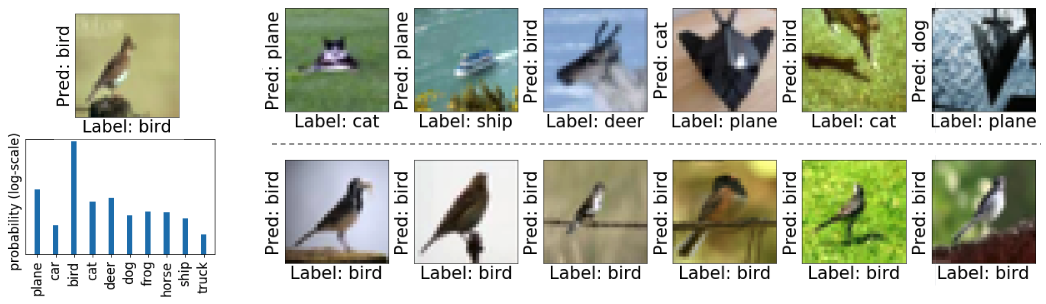}\\
	    \hline\\
	    \includegraphics[width=\linewidth]{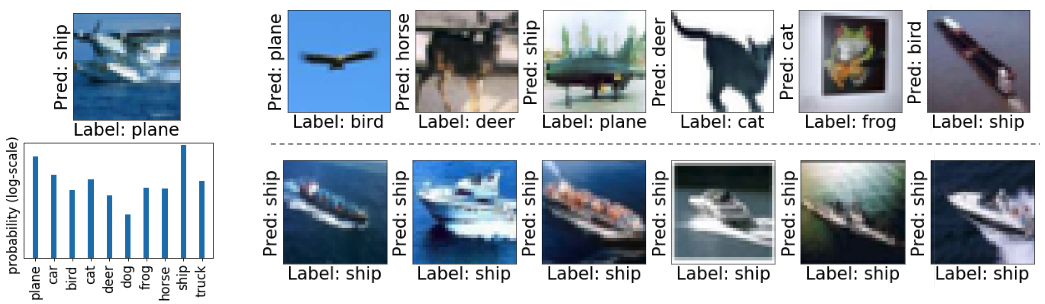}\\
	\end{tabular}
	\caption{Generating example-based explanation using IF and \RelIF{} for the model prediction. The model is a ConvNet trained on CIFAR10 data set. For each test sample in the left column, the recovered training examples for explaining the model prediction using IF and \RelIF{} are depicted in the first and second row, respectively.}
	\label{fig:exp_cifar_app1}
\end{figure*}

\begin{figure*}[h]
	\centering
	\begin{tabular}{c}
	    \includegraphics[width=\linewidth]{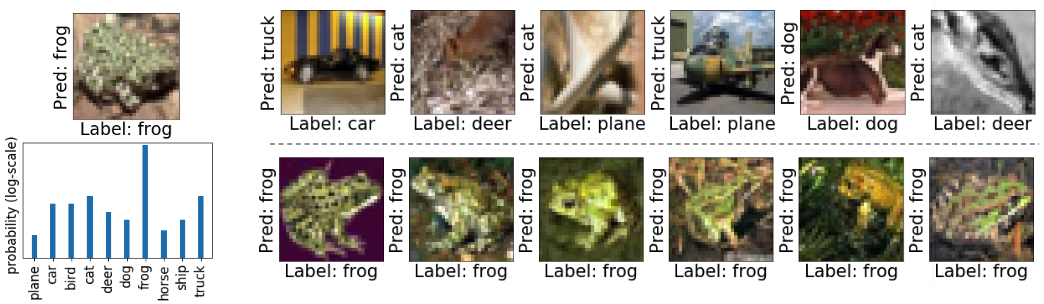}\\
	    \hline\\
	    \includegraphics[width=\linewidth]{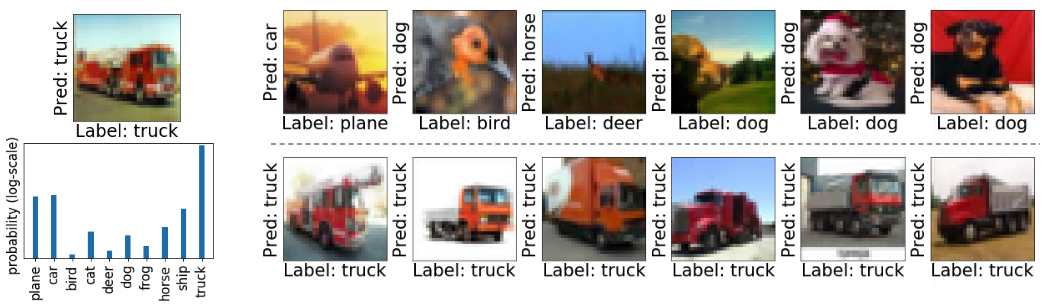}\\
	    \hline\\
	    \includegraphics[width=\linewidth]{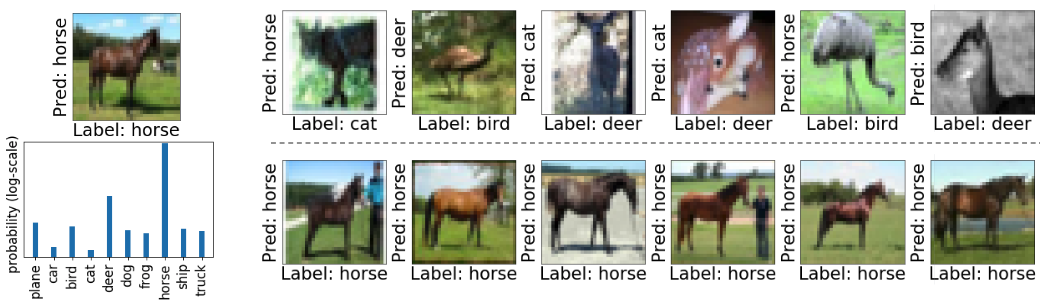}\\
	\end{tabular}
	\caption{Generating example-based explanation using IF and \RelIF{} for the model prediction. The model is a ConvNet trained on CIFAR10 data set. For each test sample in the left column, the recovered training examples for explaining the model prediction using IF and \RelIF{} are depicted in the first and second row, respectively.}
	\label{fig:exp_cifar_app2}
\end{figure*}

%----------------------------------------------------------------------
\cref{fig:exp_cifar_app1} and \cref{fig:exp_cifar_app2} offer more examples of using \RelIF{} and IF for explaining the prediction of a ConvNet trained on CIFAR10 data set.
\label{app:exp_examples}

%----------------------------------------------------------------------
\section{The connection between \RelIF{} and leave-one-out retraining}
\label{app:relatif_loo}
%----------------------------------------------------------------------
\RelIF{} is an approximation for the ratio of change in the test sample loss to global changes of the model (i.e., norm of change in the model parameters or root sum of square change in loss over the training set) resulted from leave-one-out retraining. \cref{tab:relatif_loo} reports this ratio for different methods.
\begin{table}[h]
	\centering
  %% Including Ratios
	\resizebox{\linewidth}{!}{
	\begin{tabular}{|l|c|c|}
	    \hline
		Method & $\Delta\loss_{\testpnt}/{\norm{\Delta\theta}}$  & $\Delta\loss_{\testpnt}/\sqrt{\sum(\Delta\loss_i)^2}$ \\
		\hline
		IF & $0.311\pm0.167$ & $0.016\pm0.008$ \\
		\hline
		\ThetaRelIF & $0.459\pm0.226$ & $0.026\pm0.010$ \\
		\hline
		\LossRelIF{} & $0.475 \pm0.229$ & $0.027\pm0.010$ \\
		\hline
		Nearest-N & $0.304 \pm0.182$ & $0.018\pm0.009$ \\
		\hline
	\end{tabular}
	}
	\caption{The ratio of change in the test loss ($\Delta\loss_{\testpnt}$) to the global changes, i.e., the norm of the change in the model parameters ($\norm{\Delta\theta}$), or root sum of square change in loss over the training set ($\sqrt{\sum(\Delta\loss_i)^2}$). The results come from removing the mostly positively influential training sample as determined by different methods, then retraining the model (i.e., leave-one-out retraining). The model is a logistic regression trained on MNIST. The experiment is repeated for 100 randomly selected test samples. The mean $\pm$ standard error are reported. One can approximate the leave-one-out change of these ratios using \ThetaRelIF{} and \LossRelIF{}, respectively. Note that these ratios are higher for the samples identified by RelatIF than for those identified by IF or NN. This is due to the fact that RelatIF identifies the points that maximize these ratios.}
	\label{tab:relatif_loo}
\end{table}
\clearpage
\clearpage
%%

% \begin{figure*}
% \centering
% \begin{tabular}{cc}
%     \multirow{2}{*}{\includegraphics[width=0.18\linewidth]{./figures/test_sample_1.pdf}} & \includegraphics[width=0.75\linewidth]{./figures/kNN_samples_1.pdf}\\
%      & \includegraphics[width=0.75\linewidth]{./figures/RelatIF_samples_1.pdf}\\
%      \hline\\
%      \multirow{2}{*}{\includegraphics[width=0.18\linewidth]{./figures/test_sample_2.pdf}} & \includegraphics[width=0.75\linewidth]{./figures/kNN_samples_2.pdf}\\
%      & \includegraphics[width=0.75\linewidth]{./figures/RelatIF_samples_2.pdf}\\
% \end{tabular}
% \end{figure*}
% NORMALLY END THE DOCUMENT HERE !!!!!!!!!!!!!!
\end{document}

%% file: math_commands.tex
\usepackage{amsmath,amsfonts,bm}

\def\eqref#1{equation~\ref{#1}}
\def\1{\bm{1}}

\DeclareMathAlphabet{\mathsfit}{\encodingdefault}{\sfdefault}{m}{sl}
\SetMathAlphabet{\mathsfit}{bold}{\encodingdefault}{\sfdefault}{bx}{n}

\newcommand{\E}{\mathbb{E}}

\DeclareMathOperator*{\argmax}{arg\,max}

%% file: defs.tex
\def\[#1\]{\begin{align}#1\end{align}}
\def\*[#1\]{\begin{align*}#1\end{align*}}

\newcommand\optparen[1]{\ifthenelse{\equal{#1}{}}{}{(#1)}}

\newcommand{\Reals}{\mathbb{R}}

\newcommand{\grad}{\nabla}

\newcommand{\dee}{\mathrm{d}}

\DeclareMathOperator*{\newlim}{\mathrm{lim}\vphantom{\mathrm{infsup}}}

\DeclareMathOperator*{\newmax}{\mathrm{max}\vphantom{\mathrm{infsup}}}

\renewcommand{\lim}{\newlim}

\renewcommand{\max}{\newmax}

\newcommand{\abs}[1]{\lvert #1 \rvert}
\newcommand{\norm}[1]{\lVert #1 \rVert}

\newcommand{\loss}{\ell}

\newcommand{\Infl}[2]{\mathcal{I}_{#1,#2}}

\newcommand{\testpnt}{\text{test}}

\newcommand{\gradient}[1]{g_{#1}}
\newcommand{\gradfunc}[1]{g({#1})}
\newcommand{\thetaopt}{\theta^{*}}

\newcommand{\thetaepsi}{\theta^{*}_{i,\epsilon_i}}

\newcommand{\ThetaRelIF}{$\theta$-RelatIF}
\newcommand{\LossRelIF}{$\loss$-RelatIF}

\newcommand{\RelIF}{RelatIF}

\newcommand{\inner}[2]{{\langle #1, #2 \rangle}}